\documentclass[runningheads]{llncs}
\usepackage{graphicx}

\usepackage{amsmath,amssymb} 
\usepackage{color}
\usepackage{hyperref}
\usepackage{url}
\usepackage{authblk}

\usepackage{multirow}
\usepackage{color}
\usepackage{amsfonts}
\usepackage{graphicx}
\usepackage{booktabs}
\usepackage{url}
\usepackage{perpage} 
\usepackage{tablefootnote}
\MakePerPage{footnote}
\usepackage[symbol]{footmisc}

\usepackage{paralist}
\usepackage{wrapfig}
\usepackage{enumitem}
\usepackage{comment}

\usepackage[english]{babel}
\usepackage[utf8x]{inputenc}

\def \x {\mathbf{x}}

\def \x {\mathbf{x}}

\begin{document}
\pagestyle{headings}
\mainmatter
\def\ECCV18SubNumber{3015}  









%
\title{Is Robustness the Cost of Accuracy? \\-- A Comprehensive Study on the Robustness of 18 Deep Image Classification Models}
%
%
\author{Dong Su\inst{1}\textsuperscript{*}\and
Huan Zhang\inst{2}\textsuperscript{*} \and
Hongge Chen\inst{3}\and
Jinfeng Yi\inst{4} \and   
\ \ \ \ \  \ \ \ \ \  \ \ \ \ \  \ \ \ \ \ \ \ \ \ \   Pin-Yu Chen \inst{1} \and
Yupeng Gao \inst{1}}

\authorrunning{D. Su, H. Zhang, H. Chen, J. Yi, P-Y. Chen and Y. Gao}
\titlerunning{Is Robustness the Cost of Accuracy?}
%

\institute{
   \small
   $^1$IBM Research \quad
   $^2$University of California, Davis \quad\\
   $^3$Massachusetts Institute of Technology \quad
   $^4$JD AI Research \quad
   \email{sudong.tom@gmail.com, huan@huan-zhang.com, chenhg@mit.edu, pin-yu.chen@ibm.com, yijinfeng@jd.com, yupeng.gao@ibm.com}\\
\textsuperscript{*}Dong Su and Huan Zhang contribute equally to this work
}


%
\maketitle              

\begin{abstract}
The prediction accuracy has been the long-lasting and sole standard for comparing the performance of different image classification models, including the ImageNet competition. However, recent studies have highlighted the lack of robustness in well-trained deep neural networks to adversarial examples. Visually imperceptible perturbations to natural images can easily be crafted and mislead the image classifiers towards misclassification. To demystify the trade-offs between robustness and accuracy, in this paper we thoroughly benchmark 18 ImageNet models using multiple robustness metrics, including the distortion, success rate and transferability of adversarial examples between 306 pairs of models. Our extensive experimental results reveal several new insights: (1) linear scaling law - the empirical $\ell_2$ and $\ell_\infty$ distortion metrics scale linearly with the logarithm of classification error; (2) model architecture is a more critical factor to robustness than model size, and the disclosed accuracy-robustness Pareto frontier can be used as an evaluation criterion for ImageNet model designers; (3) for a similar network architecture, increasing network depth slightly improves robustness in $\ell_\infty$ distortion;  (4) there exist models (in VGG family) that exhibit high adversarial transferability, while most adversarial examples crafted from one model can only be transferred within the same family.  Experiment code is publicly available at \url{https://github.com/huanzhang12/Adversarial_Survey}. 
 
\end{abstract}

\section{Introduction}
Image classification is a fundamental problem in computer vision and serves as the foundation of multiple tasks such as object detection, image segmentation, object tracking, action recognition, and autonomous driving. Since the breakthrough achieved by AlexNet~\cite{krizhevsky2012imagenet} in ImageNet Challenge (ILSVRC) 2012~\cite{DBLP:journals/ijcv/RussakovskyDSKS15}, deep neural networks (DNNs) have become the dominant force in this domain. 
From then on, DNN models with increasing depth and more complex building blocks have been proposed. While these models continue to achieve steadily increasing accuracies, their robustness has not been thoroughly studied, thus little is known if the high accuracies come at the price of reduced robustness. 

A common approach to evaluate the robustness of DNNs is via adversarial attacks~\cite{DBLP:journals/corr/GoodfellowSS14,xu2017can,chen2018attacking,metzen2017universal,cheng2018seq2sick,carlini2018audio,sun2018identify,ijcai2018-543,xiao2018spatially}, where imperceptible adversarial examples are crafted to mislead DNNs.  
Generally speaking, the easier an adversarial example can be generated, the less robust the DNN is. Adversarial examples may lead to significant property damage or loss of life. For example, \cite{DBLP:journals/corr/EvtimovEFKLPRS17} has shown that a subtly-modified physical Stop sign can be misidentified by a real-time object recognition system as a Speed Limit sign. In addition to adversarial attacks, neural network robustness can also be estimated in an attack-agnostic manner. For example, \cite{szegedy2013intriguing} and 
\cite{hein2017formal} theoretically analyzed the robustness of some simple neural networks by estimating their global and local Lipschitz constants, respectively. \cite{weng2018evaluating} proposes to use extreme value theory to estimate a lower bound of the minimum adversarial distortion, and can be efficiently applied to any neural network classifier. \cite{weng2018towards} proposes a robustness lower bound based on linear approximations of ReLU activations. In this work, we evaluate DNN robustness by using specific attacks as well as attack-agnostic approaches. We also note that the adversarial robustness studied in this paper is different from \cite{stock2017convnets}, where ``robustness'' is studied in the context of label semantics and accuracy.

Since the last ImageNet challenge has ended in 2017, we are now at the beginning of post-ImageNet era. In this work, we revisit 18 DNN models submitted to the ImageNet Challenge or achieved state-of-the-art performance. These models have different sizes, classification performance, and belong to multiple architecture families such as AlexNet~\cite{krizhevsky2012imagenet}, VGG Nets~\cite{simonyan2014very}, Inception Nets~\cite{DBLP:conf/cvpr/SzegedyLJSRAEVR15}, ResNets~\cite{DBLP:conf/cvpr/HeZRS16}, DenseNets~\cite{DBLP:journals/corr/HuangLW16a}, MobileNets~\cite{DBLP:journals/corr/HowardZCKWWAA17}, and NASNets~\cite{zoph2017learning}. Therefore, they are suitable to analyze how different factors influence the model robustness. Specifically, we aim to examine
the following questions in this study:
\begin{enumerate}
\item \emph{Has robustness been sacrificed for the increased classification performance?}
\item \emph{Which factors influence the robustness of DNNs?}
\end{enumerate}

In the course of evaluation, we have gained a number of insights and we summarize our contributions as follows:
\begin{itemize}
[nosep,wide,labelindent=0pt,labelwidth=*,align=left]
\item Tested on a large number of well-trained deep image classifiers, we find that robustness is scarified when solely pursuing a higher classification performance. Indeed, Figure~\ref{fig:accuracy-robustness}(a) and Figure~\ref{fig:accuracy-robustness}(b) clearly show that the $\ell_2$ and $\ell_\infty$ adversarial distortions scale almost linearly with the logarithm of model classification errors. 
Therefore, the classifiers with very low test errors are highly vulnerable to adversarial attacks. We advocate that ImageNet network designers should evaluate model robustness via our disclosed accuracy-robustness Pareto frontier.


\item The networks of a same family, e.g., VGG, Inception Nets, ResNets, and DenseNets, share similar robustness properties. This suggests that network architecture has a larger impact on robustness than model size.
Besides, we also observe that the $\ell_\infty$ robustness slightly improves when ResNets, Inception Nets, and DenseNets become deeper. 


\item The adversarial examples generated by the VGG family can transfer very well to all the other 17 models, while most adversarial examples of other models can only transfer within the same model family. Interestingly, this finding provides us an opportunity to reverse-engineer the architecture of black-box models.

\item We present the first comprehensive study that compares the robustness of 18 popular and state-of-the-art ImageNet models, offering a complete picture of the accuracy v.s. robustness trade-off. In terms of transferability of adversarial examples, we conduct thorough experiments on each pair of the 18 ImageNet networks (306 pairs in total), which is the largest scale to date.

\end{itemize}

\section{Background and Experimental Setup}
In this section, we introduce the background knowledge and how we set up experiments. We study both untargeted attack and targeted attack in this paper. Let $\x_0$ denote the original image and $\x$ denote the adversarial image of $\x_0$. The DNN model $F(\cdot)$ outputs a class label (or a probability distribution of class labels) as the prediction. Without loss of generality, we assume that $F(\x_0)=y_0$, which is the ground truth label of $\x_0$, to avoid trivial solution. For untargeted attack, the adversarial image $\x$ is crafted in a way that $\x$ is close to $\x_0$ but $F(\x)\neq y_0$. For targeted attack, a target class $t$ ($t\neq y_0$) is provided and the adversarial image $\x$ should satisfy that (i) $\x$ is close to $\x_0$, and (ii) $F(\x)=t$. 


\subsection{Deep Neural Network Architectures}

In this work, we study the robustness of 18 deep image classification models belonging to 7 architecture families, as summarized below. Their basic properties of these models are given in Table \ref{tab:models}.
\begin{itemize}
[nosep,wide,labelindent=0pt,labelwidth=*,align=left]
\item \textbf{AlexNet} AlexNet~\cite{krizhevsky2012imagenet} is one of the pioneering and most well-known deep convolutional neural networks. 
Compared to many recent architectures, AlexNet has a relatively simple layout that is composed of $5$ convolutional layers followed by two fully connected layers and a softmax output layer. 
\item \textbf{VGG Nets} The overall architecture of VGG nets~\cite{simonyan2014very} are similar to AlexNet, but they are much deeper with more convolutional layers. Another main difference between VGG nets and AlexNet is that all the convolutional layers of VGG nets use a small (3$\times$3) kernel while the first two layers of AlexNet use 11$\times$11 and 5$\times$5 kernels, respectively. 
In our paper, we study VGG networks with $16$ and $19$ layers, with 138 million and 144 million parameters, respectively. 
\item \textbf{Inception Nets} 
The family of Inception nets utilizes the inception modules~\cite{lin2013network} that act as multi-level feature extractors. Specifically, each inception module consists of multiple branches of $1\times 1$, $3\times 3$, and $5\times 5$ filters, whose outputs will stack along the channel dimension and be fed into the next layer in the network. In this paper, we study the performance of all popular networks in this family, including Inception-v1 (GoogLeNet)~\cite{DBLP:conf/cvpr/SzegedyLJSRAEVR15}, Inception-v2~\cite{DBLP:conf/icml/IoffeS15}, Inception-v3~\cite{DBLP:conf/cvpr/SzegedyVISW16}, Inception-v4, and Inception-ResNet~\cite{DBLP:conf/aaai/SzegedyIVA17}. All these models are much deeper than AlexNet/VGG but have significantly fewer parameters.

\item \textbf{ResNets} 
To solve the vanishing gradient problem for training very deep neural networks, the authors of \cite{DBLP:conf/cvpr/HeZRS16} proposes ResNets, where each layer learns the residual functions with reference to the input by adding skip-layer paths, or ``identity shortcut connections''. This architecture enables practitioners to train very deep neural networks to outperform shallow models. In our study, we evaluate 3 ResNets with different depths. 
\item \textbf{DenseNets} To further exploit the ``identity shortcut connections" techniques from ResNets, \cite{DBLP:journals/corr/HuangLW16a} proposes DenseNets that connect all layers with each other within a dense block. Besides tackling gradient vanishing problem, the authors also claimed other advantages such as encouraging feature reuse and reducing the number of parameters in the model. We study 3 DenseNets with different depths and widths.  
\item \textbf{MobileNets} MobileNets~\cite{DBLP:journals/corr/HowardZCKWWAA17} are a family of light weight and efficient neural networks designed for mobile and embedded systems with restricted computational resources. The core components of MobileNets are depthwise separable filters with factorized convolutions. Separable filters can factorize a standard convolution into two parts, a depthwise convolution and a $1\times 1$ pointwise convolution, which can reduce computation and model size dramatically. In this study, we include 3 MobileNets with different depths and width multipliers.
\item \textbf{NASNets} NASNets~\cite{zoph2017learning} are a family of networks automatically generated by reinforcement learning using a policy gradient algorithm to optimize architectures~\cite{zoph2016neural}. Building blocks of the model are first searched on a smaller dataset and then transfered to a larger dataset.
\end{itemize}

\begin{table*}[t]
\begin{minipage}{\textwidth}
   \caption{18 ImageNet models under robustness examination}
    \label{tab:models}
    \centering
    
  \begin{tabular}{|l|c|c|c|c|}
      \hline
   Models \ \  & \   Year \   & \  \# layers\   & \   \# parameters \   & \  Top-1/5 ImageNet accuracies\   \\
      \hline\hline
   AlexNet~\cite{krizhevsky2012imagenet}& 2012& 8 &60 million & \!\!\!\!\!\!\!56.9\% /	80.1\%~\footnote[1]{\scriptsize{\url{https://github.com/BVLC/caffe/wiki/Models-accuracy-on-ImageNet-2012-val}} }\\
   \hline
   VGG~16~\cite{simonyan2014very} & 2014&  16& 138 million& 71.5\% /	89.8\%\cite{tensorflow_models}\\
    \cline{1-5}
   VGG~19~\cite{simonyan2014very}  & 2014&  19& 144 million&71.1\% /	89.8\%\cite{tensorflow_models}\\
   \hline
   Inception-v1~\cite{DBLP:conf/cvpr/SzegedyLJSRAEVR15} & 2014& 22 & 6.7 million& 69.8\% /	89.6\%\cite{tensorflow_models}\\
   \hline
   Inception-v2~\cite{DBLP:conf/icml/IoffeS15} & 2015& 48 & 11.3 million& 73.9\% / 91.8\%\cite{tensorflow_models}\\
   \hline
   Inception-v3~\cite{DBLP:conf/cvpr/SzegedyVISW16} & 2015& 48 & 23.9 million& 78.0\% /	93.9\%\cite{tensorflow_models}\\
   \hline
   Inception-v4~\cite{DBLP:conf/aaai/SzegedyIVA17} & 2016& 76 & 42.9 million& 80.2\% / 95.2\%\cite{tensorflow_models}\\
   \hline
   Inception-ResNet-v2~\cite{DBLP:conf/aaai/SzegedyIVA17} &2016&96&56.1 million&80.4\% / 95.3\%\cite{tensorflow_models}\\
   \hline
   ResNet-v2-50~\cite{he2016identity} & 2016& 50 & 25.7 million& 75.6\% / 92.8\%\cite{tensorflow_models}\\
   \hline
   ResNet-v2-101~\cite{he2016identity} & 2016& 101 & 44.8 million& 77.0\% / 93.7\%\cite{tensorflow_models}\\
   \hline
   ResNet-v2-152~\cite{he2016identity} & 2016& 152 & 60.6 million& 77.8\% / 94.1\%\cite{tensorflow_models}\\
   \hline
   DenseNet-121-k32~\cite{DBLP:journals/corr/HuangLW16a}& 2017& 121 &8.2 million & \!\!\!\!\!\!74.9\% / 92.2
\%~\footnote[2]{\scriptsize{\url{https://github.com/pudae/tensorflow-densenet}}\label{note2}} \\
   \hline
   DenseNet-169-k32~\cite{DBLP:journals/corr/HuangLW16a} & 2017& 169 & 14.4 million& \!\!\!\!\!\!76.1\% / 93.1
  \%~\textsuperscript{\ref{note2}}\\
   \hline
   DenseNet-161-k48~\cite{DBLP:journals/corr/HuangLW16a} & 2017& 161 & 29.0 million& \!\!\!\!\!\!77.6\% / 93.8
  \%~\textsuperscript{\ref{note2}}\\
   \hline
   MobileNet-0.25-128~\cite{DBLP:journals/corr/HowardZCKWWAA17} & 2017& 128 & 0.5 million& 41.5\% / 66.3\%\cite{tensorflow_models}\\
   
   \hline
   MobileNet-0.50-160~\cite{DBLP:journals/corr/HowardZCKWWAA17} & 2017&160  & 1.4 million& 59.1\% / 81.9\%\cite{tensorflow_models}\\
   \hline
   MobileNet-1.0-224~\cite{DBLP:journals/corr/HowardZCKWWAA17} & 2017& 224 & 4.3 million&70.9\% / 89.9\% \cite{tensorflow_models}\\
   \hline
   NASNet~\cite{zoph2017learning} & 2017& - &88.9 million&82.7\% / 96.2\%\cite{tensorflow_models}\\
   \hline
  \end{tabular}
  \end{minipage}
\end{table*}

\subsection{Robustness Evaluation Approaches}
We use both adversarial attacks and attack-agnostic approaches to evaluate network robustness. We first generate adversarial examples of each network using 
multiple state-of-the-art attack algorithms, and then analyze the attack success rates and the distortions of adversarial images. In this experiment, we assume to have full access to the targeted DNNs, known as the white-box attack. 
To further study the transferability of the adversarial images generated by each network, we consider all the 306 network pairs and for each pair, we conduct transfer attack that uses one model's adversarial examples to attack the other model. Since transfer attack is widely used in the black-box setting~\cite{liu2016delving,papernot2016transferability,chen2017zoo,DBLP:journals/corr/abs-1805-11770,cheng2018query,tu2018autozoom}, where an adversary has no access to the explicit knowledge of the target models, this experiment can provide some evidence on networks' black-box robustness. 
Finally, we compute CLEVER~\cite{weng2018evaluating} score, a state-of-the-art attack-agnostic network robustness metric, to estimate each network's intrinsic robustness. Below, we briefly introduce all the evaluation approaches used in our study.

We evaluate the robustness of DNNs using the following adversarial attacks:

\begin{itemize}[nosep,wide,labelindent=0pt,labelwidth=*,align=left]
\item \textbf{Fast Gradient Sign Method (FGSM)} FGSM~\cite{DBLP:journals/corr/GoodfellowSS14} is one of the pioneering and most efficient attacking algorithms. It only needs to compute the gradient once to generate an adversarial example $\x$: 
    \[\x\leftarrow \mathrm{clip}[\x_0-\epsilon\ \mathrm{\textbf{sgn}}(\nabla J(\x_0,t))],\]
    where $\mathrm{\textbf{sgn}}(\nabla J(\x_0,t))$ is the sign of the gradient of the training loss with respect to $\x_0$, and $\mathrm{clip}(\x)$ ensures that $\x$ stays within the range of pixel values. It is efficient for generating adversarial examples as it is just an one-step attack.
\item \textbf{Iterative FGSM (I-FGSM)} Albeit efficient, FGSM suffers from a relatively low attack success rate. To this end,~\cite{DBLP:journals/corr/KurakinGB16a} proposes iterative FGSM to enhance its performance. It applies FGSM multiple times with a finer distortion, and is able to fool the network in more than $99\%$ cases. When we run I-FGSM for $T$ iterations, we set the per-iteration perturbation to $\frac{\epsilon}{T}\ \mathrm{\textbf{sgn}}(\nabla J(\x_0,t))$. I-FGSM can be viewed as a projected gradient descent (PGD) method inside an $\ell_\infty$ ball~\cite{cisse2017parseval}, and it usually finds adversarial examples with small $\ell_\infty$ distortions.

\item \textbf{C\&W attack} \cite{DBLP:conf/sp/Carlini017} formulates the problem of generating adversarial examples $\x$ as the following optimization problem
    \begin{eqnarray}
    &\min\limits_{\x}&\lambda f(\x,t)+\|\x-\x_0\|_2^2\nonumber\\
    &\mathrm{s.t.}& \x\in [0,1]^p,\nonumber
    \end{eqnarray}
    where $f(\x,t)$ is a loss function to measure the distance between the prediction of $\x$ and the target label $t$. In this work, we choose $$f(\x,t)=\max\{\max\limits_{i\neq t}[(\mathrm{\textbf{Logit}}(\x))_i-(\mathrm{\textbf{Logit}}(\x))_t],-\kappa\}$$ as it was shown to be effective by \cite{DBLP:conf/sp/Carlini017}. $\mathrm{\textbf{Logit}}(\x)$ denotes the vector representation of $\x$ at the logit layer, $\kappa$ is a confidence level and a larger $\kappa$ generally improves transferability of adversarial examples.
    
C\&W attack is by far one of the strongest attacks that finds adversarial examples with small $\ell_2$ perturbations. It can achieve almost $100\%$ attack success rate and has bypassed $10$ different adversary detection methods~\cite{DBLP:journals/corr/CarliniW17}. 
    
\item \textbf{EAD-L1 attack} EAD-L1 attack~\cite{chen2017ead} refers to the \textbf{E}lastic-Net \textbf{A}ttacks to \textbf{D}NNs, which is a more general formulation than C\&W attack. It proposes to use elastic-net regularization, a linear combination of $\ell_1$ and $\ell_2$ norms, to penalize large distortion between the original and adversarial examples. Specifically, it learns the adversarial example $\x$ via
    \begin{eqnarray}
    &\min\limits_{\x}&\lambda f(\x,t)+\|\x-\x_0\|_2^2+\beta\|\x-\x_0\|_1\nonumber\\
    &\mathrm{s.t.}& \x\in[0,1]^p,\nonumber
    \end{eqnarray}
    where $f(\x,t)$ is 
    the same as used in the C\&W attack.  \cite{chen2017ead,sharma2017breaking,lu2018limitation,lu2018limitation2} show that EAD-L1 attack is highly transferable and can bypass many defenses and analysis.
\end{itemize}


We also evaluate network robustness using an attack-agnostic approach:

\begin{itemize}
[nosep,wide,labelindent=0pt,labelwidth=*,align=left]
\item \textbf{CLEVER} CLEVER~\cite{weng2018evaluating} (Cross-Lipschitz Extreme Value for nEtwork Robustness) uses extreme value theory to estimate a lower bound of the minimum adversarial distortion. Given an image $\x_0$, CLEVER provides an estimated lower bound on the $\ell_p$ norm of the minimum distortion $\delta$ required to misclassify the distorted image $\x_0+\delta$. A higher CLEVER score suggests that the network is likely to be more robust to adversarial examples. CLEVER is attack-agnostic and reflects the intrinsic robustness of a network, rather than the robustness under a certain attack. 
\end{itemize}

\subsection{Dataset}
In this work, we use the ImageNet~\cite{deng2009imagenet} as the benchmark dataset, due to the following reasons: (i) ImageNet dataset can take full advantage of the studied DNN models since all of them were designed for ImageNet challenges; (ii) comparing to the widely-used small-scale datasets such as MNIST, CIFAR-10~\cite{krizhevsky2009learning}, and GTSRB~\cite{DBLP:journals/nn/StallkampSSI12}, ImageNet has significantly more images and classes and is more challenging; and (iii) it has been shown by \cite{DBLP:conf/sp/Carlini017,DBLP:conf/cvpr/Moosavi-Dezfooli16} that ImageNet images are easier to attack but harder to defend than the images from 
 MNIST and CIFAR datasets. Given all these observations, ImageNet is an ideal candidate to study the robustness of state-of-the-art deep image classification models. 

A set of randomly selected 1,000 images from the ImageNet validation set is used to generate adversarial examples from each model. For each image, we conduct targeted attacks with a random target and a least likely target as well as an untargeted attack. Misclassified images are excluded. We follow the setting in~\cite{weng2018evaluating} to compute CLEVER scores for 100 out of the all 1,000 images, as CLEVER is relatively more computational expensive. Additionally, we conducted another experiment by taking the subset of images (327 images in total) that are correctly classified by \textit{all} of 18 examined ImageNet models. The results are consistent with our main results and are given in supplementary material.

\subsection{Evaluation Metrics}
In our study, the robustness of the DNN models is evaluated using the following four metrics:
\begin{itemize}[nosep,wide,labelindent=0pt,labelwidth=*,align=left]
\item \textbf{Attack success rate} For non-targeted attack, success rate indicates the percentage of the adversarial examples whose predicted labels are different from their ground truth labels. For targeted attack, success rate indicates the percentage of the adversarial examples that are classified as the target class. For both attacks, a higher success rate suggests that the model is easier to attack and hence less robust.
When generating adversarial examples, we only consider original images that are correctly classified 
to avoid trial attacks.

\item \textbf{Distortion} We measure the distortion between adversarial images and the original ones using $\ell_2$ and $\ell_\infty$ norms. $\ell_2$ norm measures the Euclidean distance between two images, and $\ell_\infty$ norm is a measure of the maximum absolute change to any pixel (worst case). Both of them are widely used to measure adversarial perturbations~\cite{DBLP:journals/corr/CarliniW17,DBLP:conf/sp/Carlini017,chen2017ead}. A higher distortion usually suggests a more robust model. 
To find adversarial examples with minimum distortion for each model, 
we use a binary search strategy to select the optimal attack parameters $\epsilon$ in I-FGSM and $\lambda$ in C\&W attack. Because each model may have different input sizes, we divide $\ell_2$ distortions by the number of total pixels for a fair comparison.

 \item \textbf{CLEVER score} For each image, we compute its $\ell_2$ CLEVER score for target attacks with a random target class and a least-likely class, respectively. The reported number is the averaged score of all the tested images. The higher the CLEVER score, the more robust the model is. 

\item \textbf{Transferability}  We follow~\cite{liu2016delving} to define targeted and non-targeted transferability. For non-targeted attack, transferability is defined as the percentage of the adversarial examples generated for one model (\emph{source model}) that are also misclassified by another model (\emph{target model}). We refer to this percentage as \emph{error rate}, and a higher error rate means better non-targeted transferability. For targeted attack, transferability is defined as \emph{matching rate}, i.e., the percentage of the adversarial examples generated for source model that are misclassified as the target label (or within top-k labels) by the target model. A higher matching rate indicates better targeted transferability. 
\end{itemize}

\section{Experiments}

After examining all the 18 DNN models, we have learned insights about the relationships between model architectures and robustness, as discussed below.

\subsection{Evaluation of Adversarial Attacks}
\begin{figure*}[!htb]
\begin{tabular}{cc}
    \includegraphics[width = 2.4in]{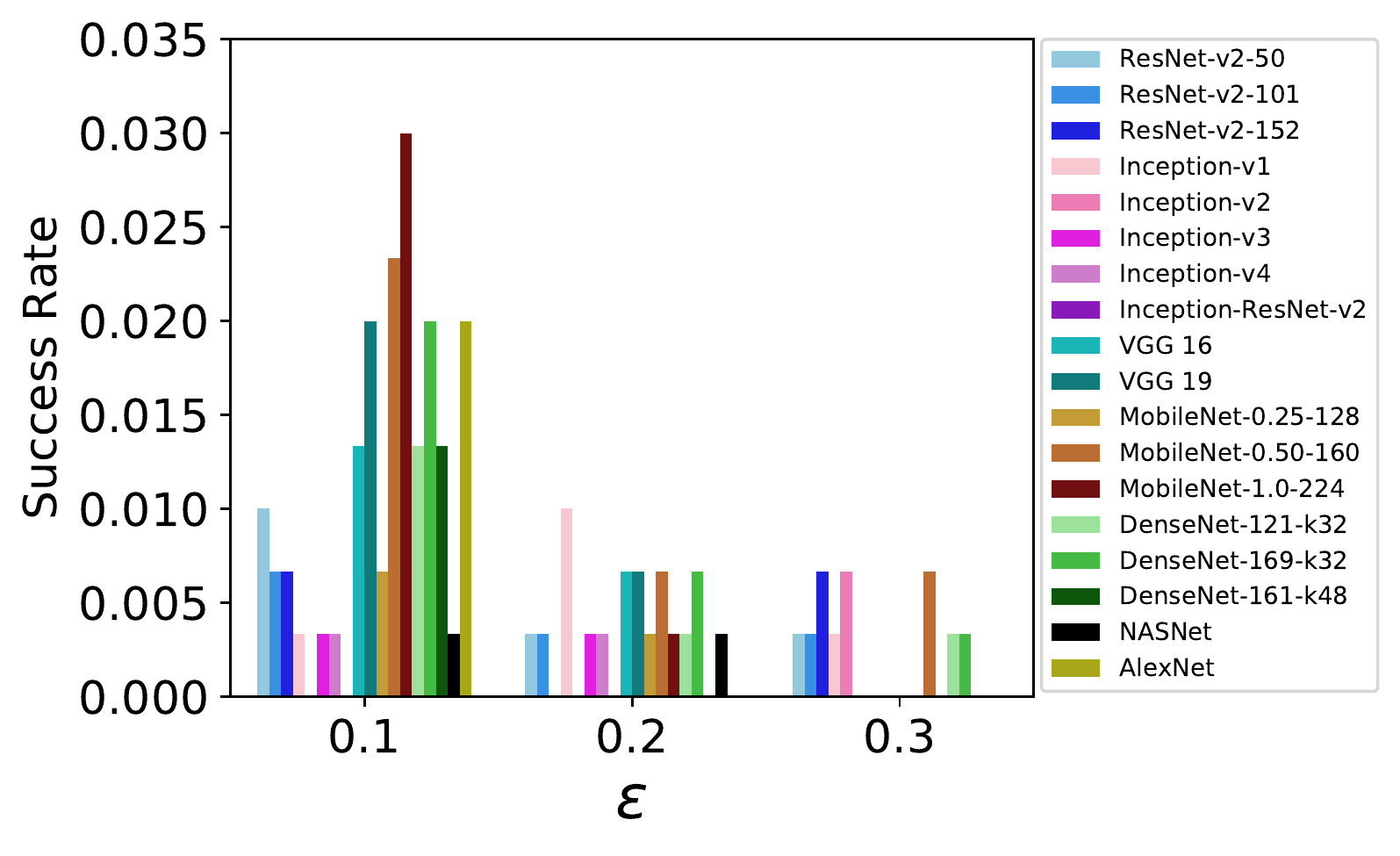}
    & \includegraphics[width = 2.4in]{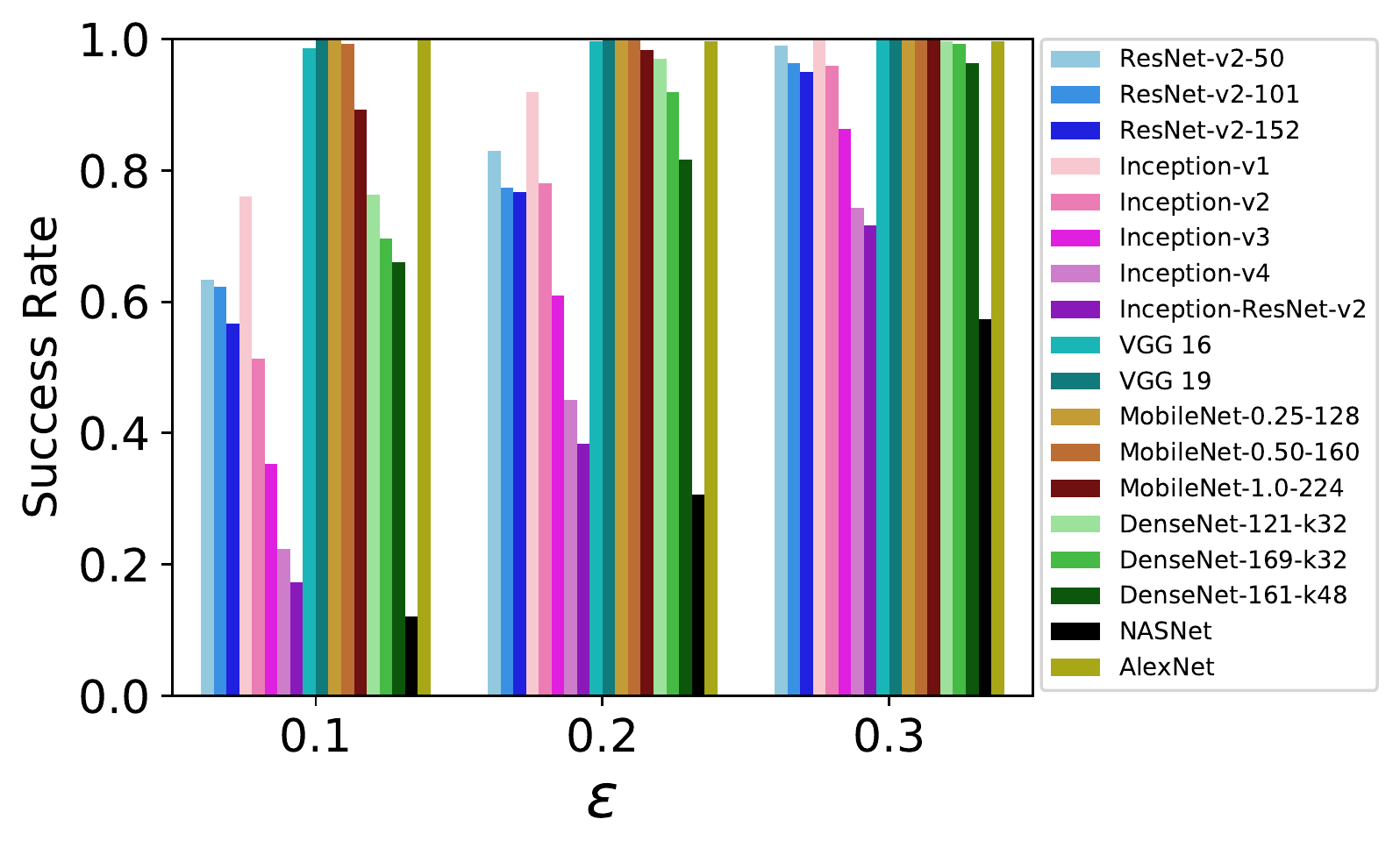}\\
    (a) Success rate, targeted FGSM & (b) Success rate, untargeted FGSM\\
    \includegraphics[width = 2.4in]{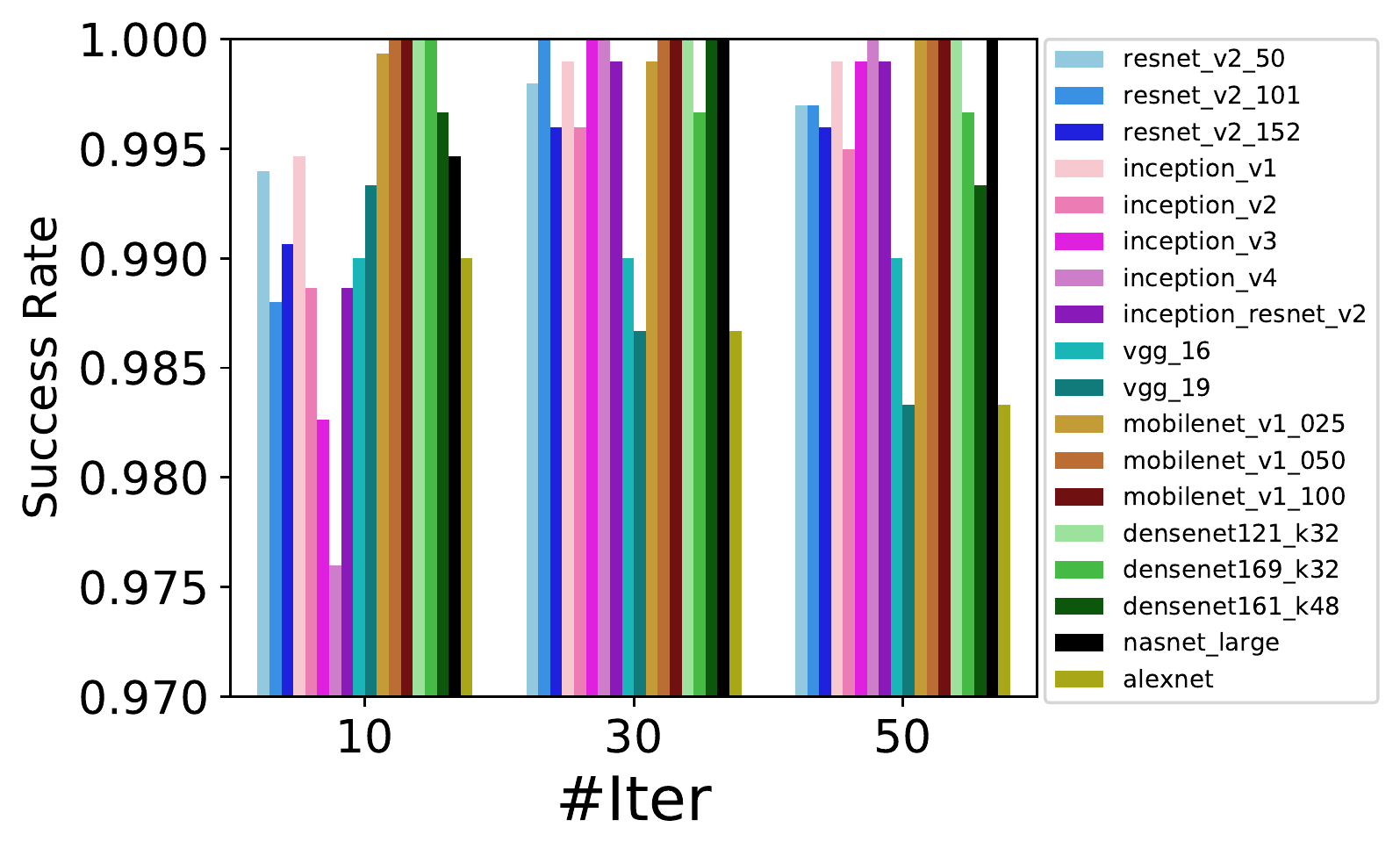}
    & \includegraphics[width = 2.4in]{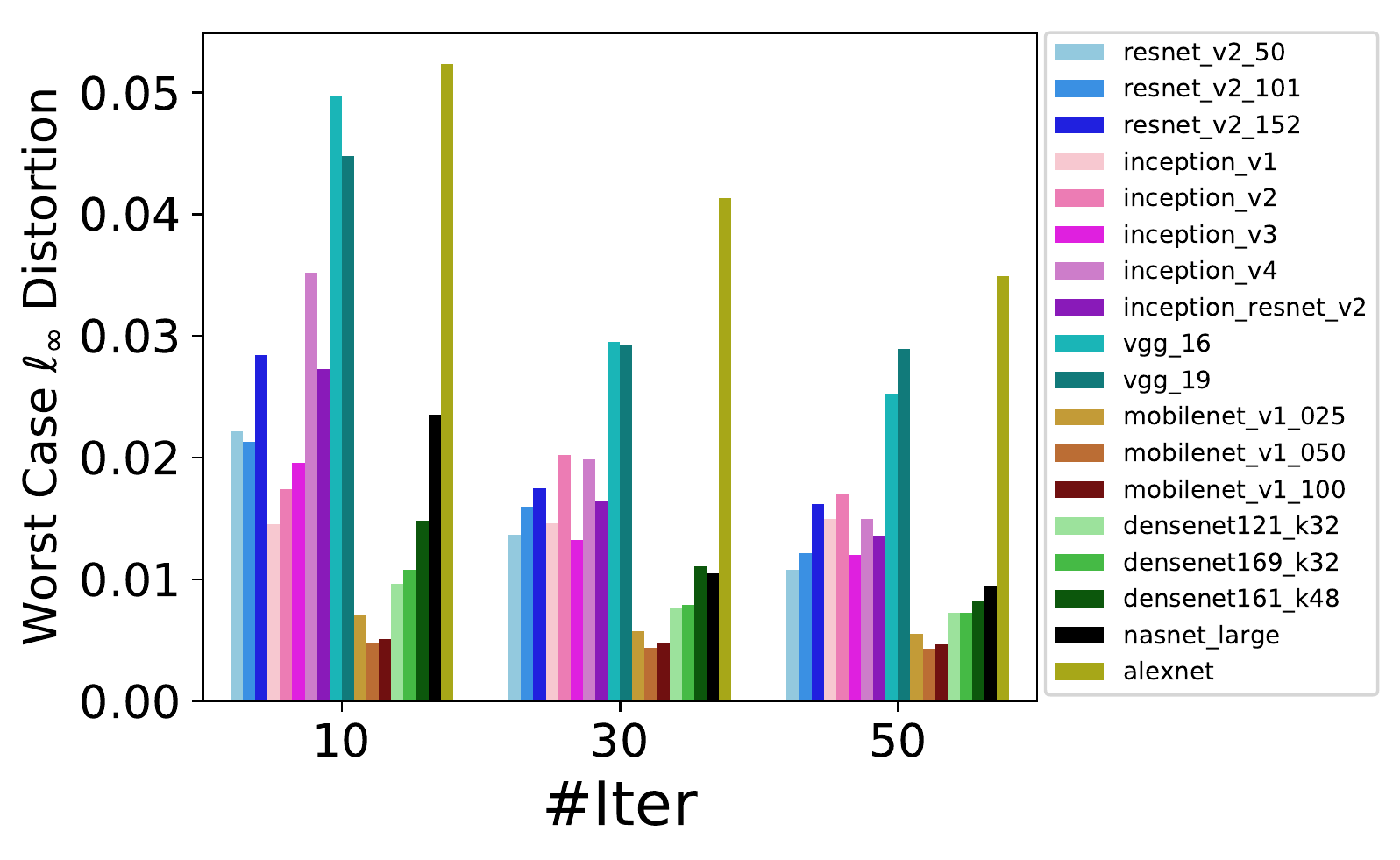}\\
    (c) Success rate, targeted I-FGSM & (d) Worst case $\ell_{\infty}$ distortion, I-FGSM, \\
    \includegraphics[width = 2.4in]{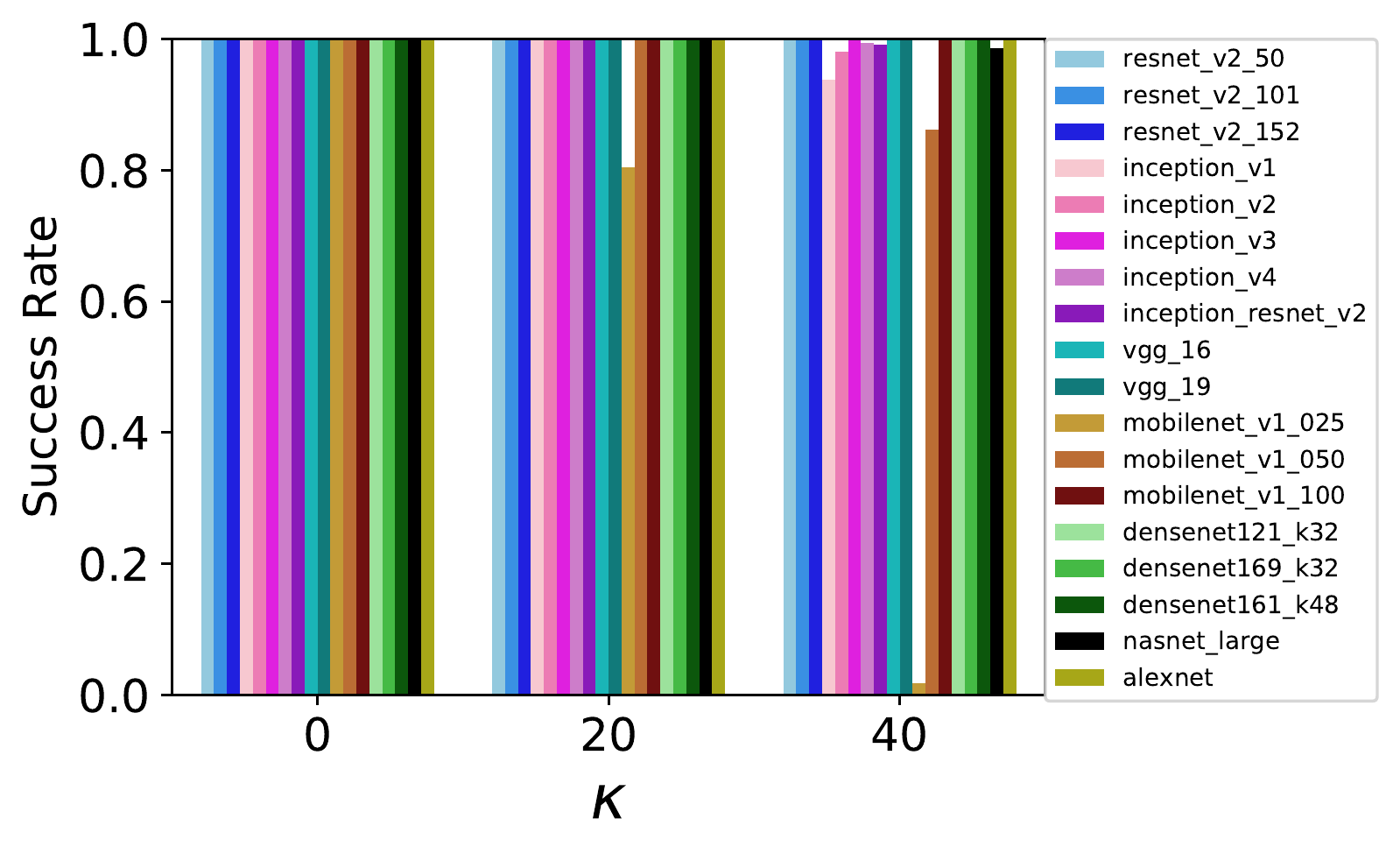}
    & \includegraphics[width = 2.4in]{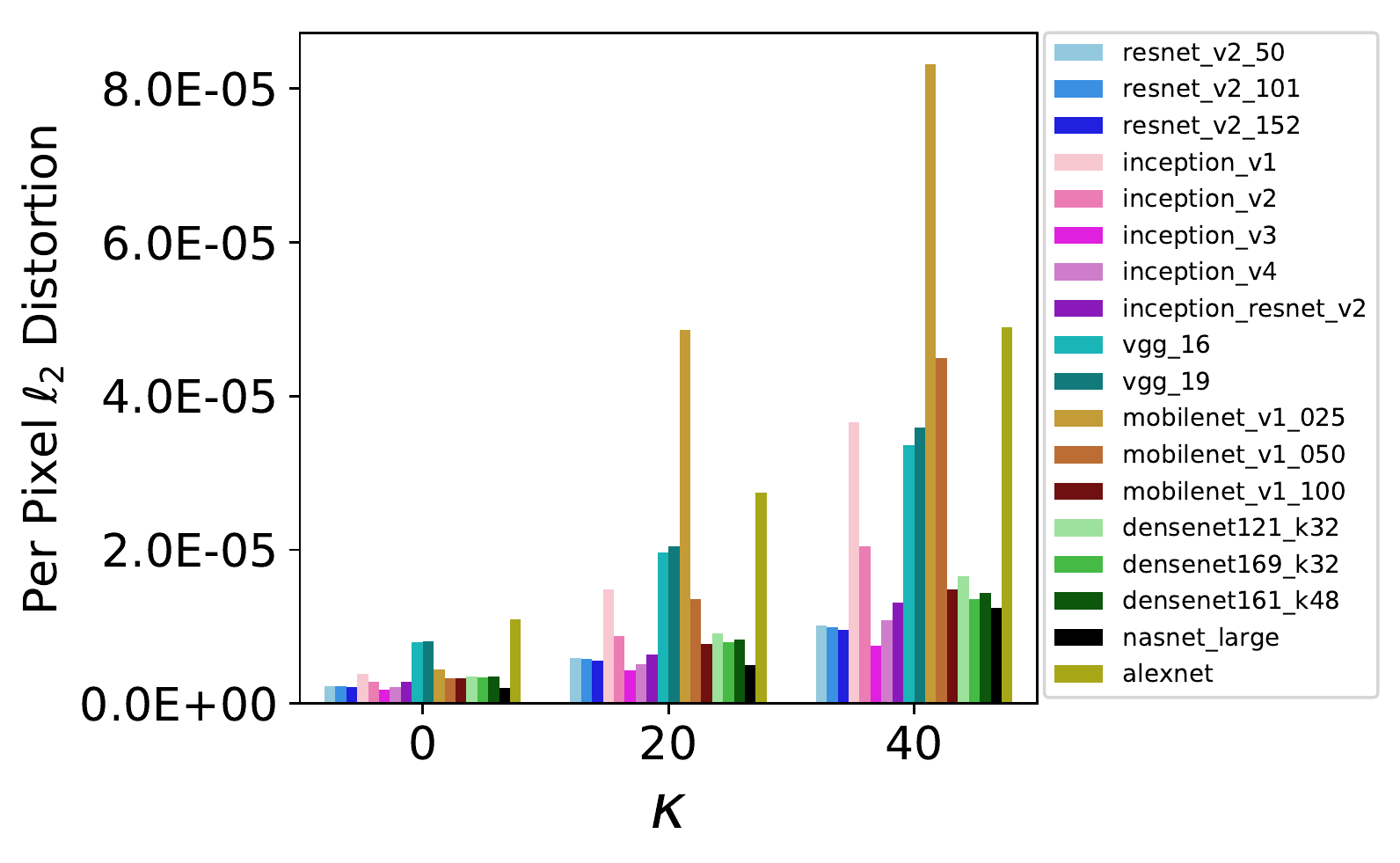} \\
    (e)  Success rate, targeted C\&W & (f) Per pixel $\ell_2$ distortion, targeted C\&W\\
    \includegraphics[width = 2.4in]{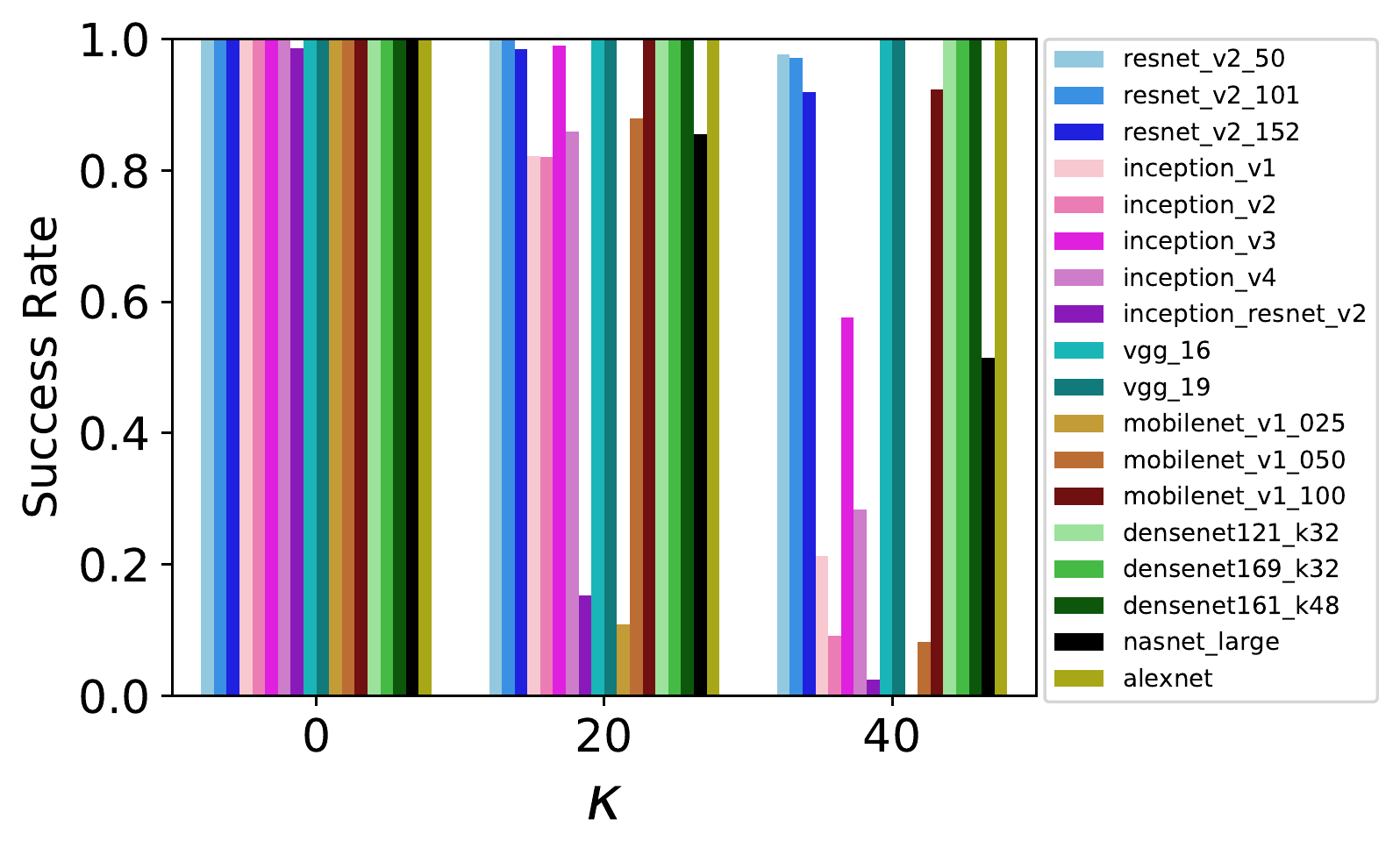}
    & \includegraphics[width = 2.4in]{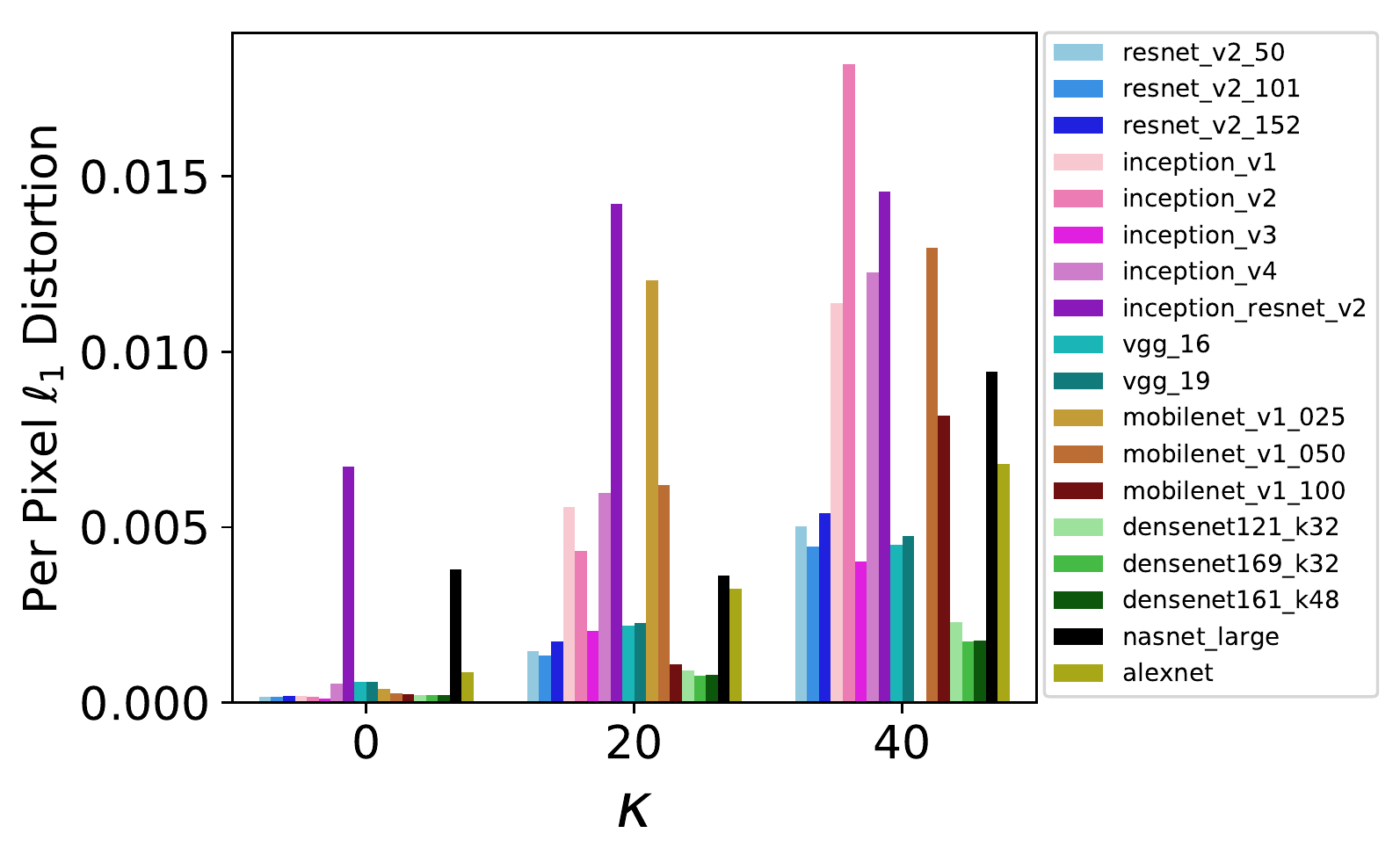} \\
    (g) Success rate, targeted EAD-L1 & (h) Per pixel $\ell_1$ distortion, targeted EAD-L1\\

\end{tabular}
    \caption{Comparison of FGSM, I-FGSM, CW and EAD-L1 attacks by varying attack parameters.}\label{fig:comparison-vary-param}
\end{figure*}

We have carefully conducted a controlled experiment by pulling images from a \textit{common} set of 1000 test images when evaluating the robustness of different models. For assessing the robustness of each model, the originally misclassified images are excluded. We compare the success rates of targeted attack with a \textit{random} target of FGSM, I-FGSM, C\&W and EAD-L1 with different parameters for all 18 models. The success rate of FGSM targeted attack is low so we also show its untargeted attack success rate in Figure~\ref{fig:comparison-vary-param}(b).

For targeted attack, the success rate of FGSM is very low (below 3\% for all settings), and unlike in the untargeted setting, increasing $\epsilon$ in fact \textit{decreases} attack success rate. This observation further confirms that FGSM is a weak attack, and targeted attack is more difficult and needs iterative attacking methods. Figure~\ref{fig:comparison-vary-param}(c) shows that, with \textit{only 10} iterations, I-FGSM can achieve a very good targeted attack success rate on all models.  C\&W and EAD-L1 can also achieve almost 100\% success rate on almost all of the models when $\kappa = 0$.  

\begin{figure*}
\centering
\begin{tabular}{c}
	\includegraphics[width = 0.8\textwidth]{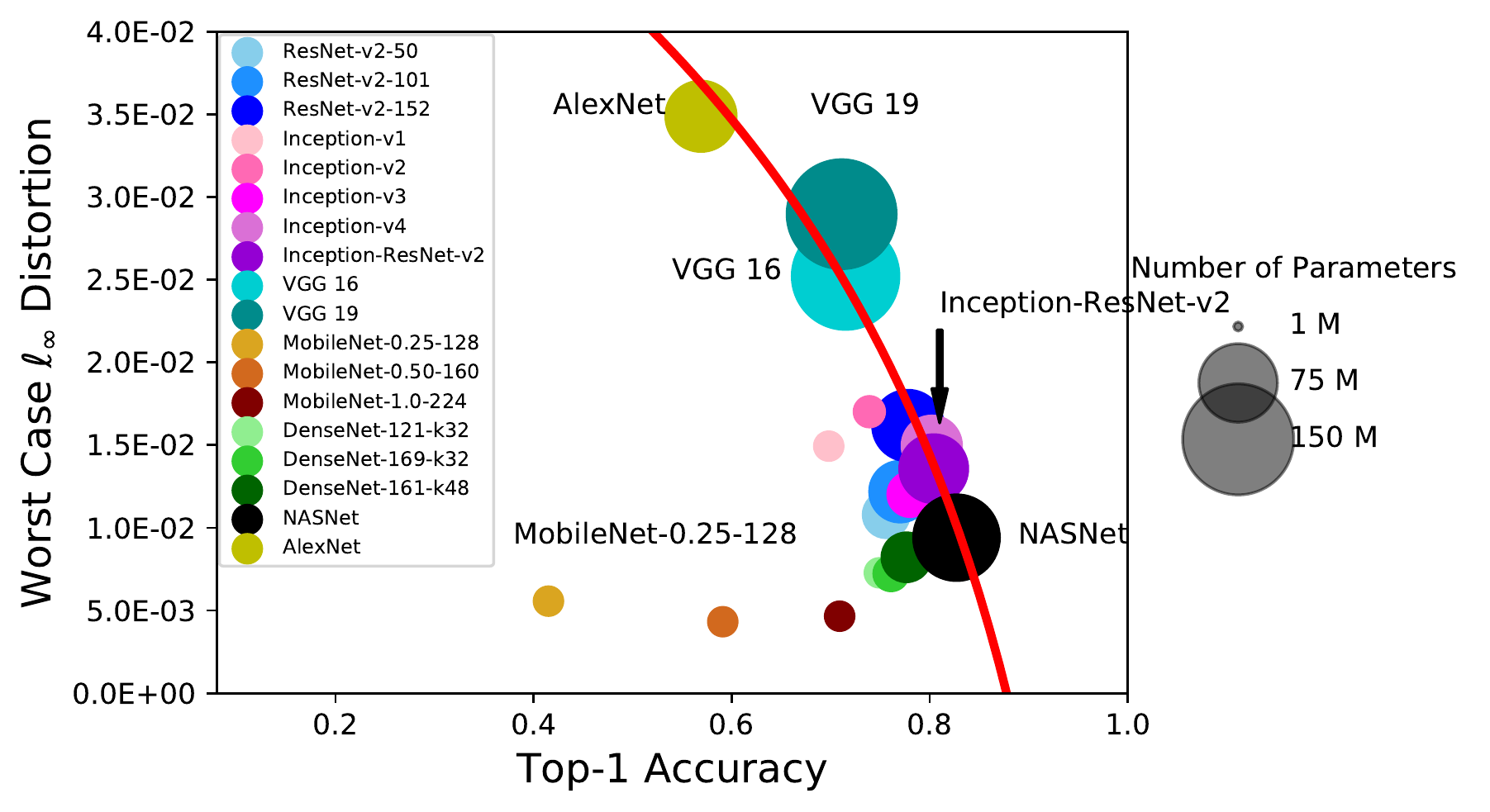}\\
	(a) Fitted Pareto frontier of $\ell_\infty$ distortion (I-FGSM attack) vs. top-1 accuracy: \\$\ell_\infty\text{ dist}=[2.9\cdot\ln(1-\text{acc})+6.2]\times 10^{-2}$\\
    \includegraphics[width = 0.8\textwidth]{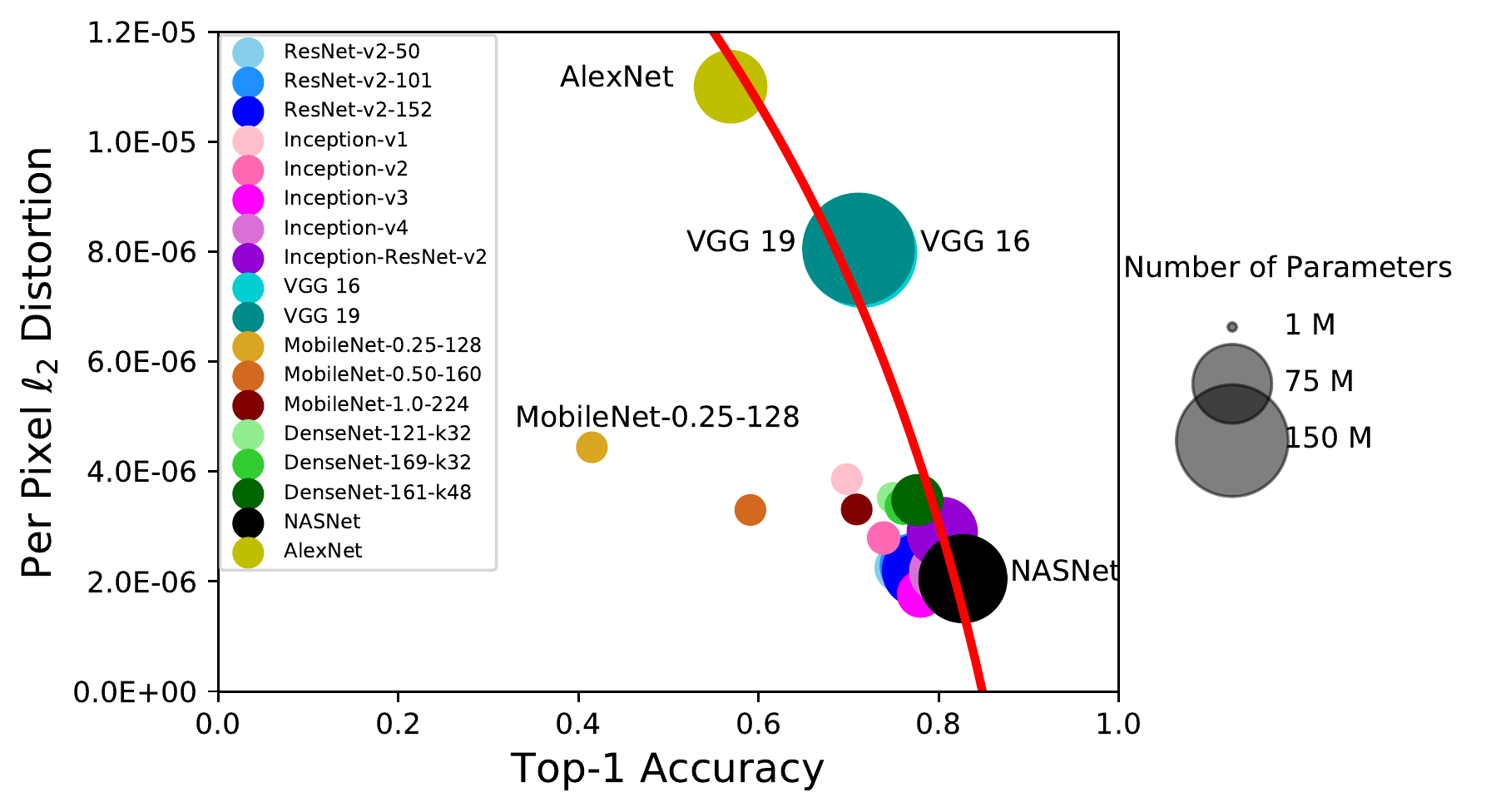}\\
	(b) Fitted Pareto frontier of $\ell_2$ distortion (C\&W attack) vs. top-1 accuracy:\\ $\ell_2\text{ dist}=[1.1\cdot\ln(1-\text{acc})+2.1]\times 10^{-5}$\\
    \includegraphics[width = 0.8\textwidth]{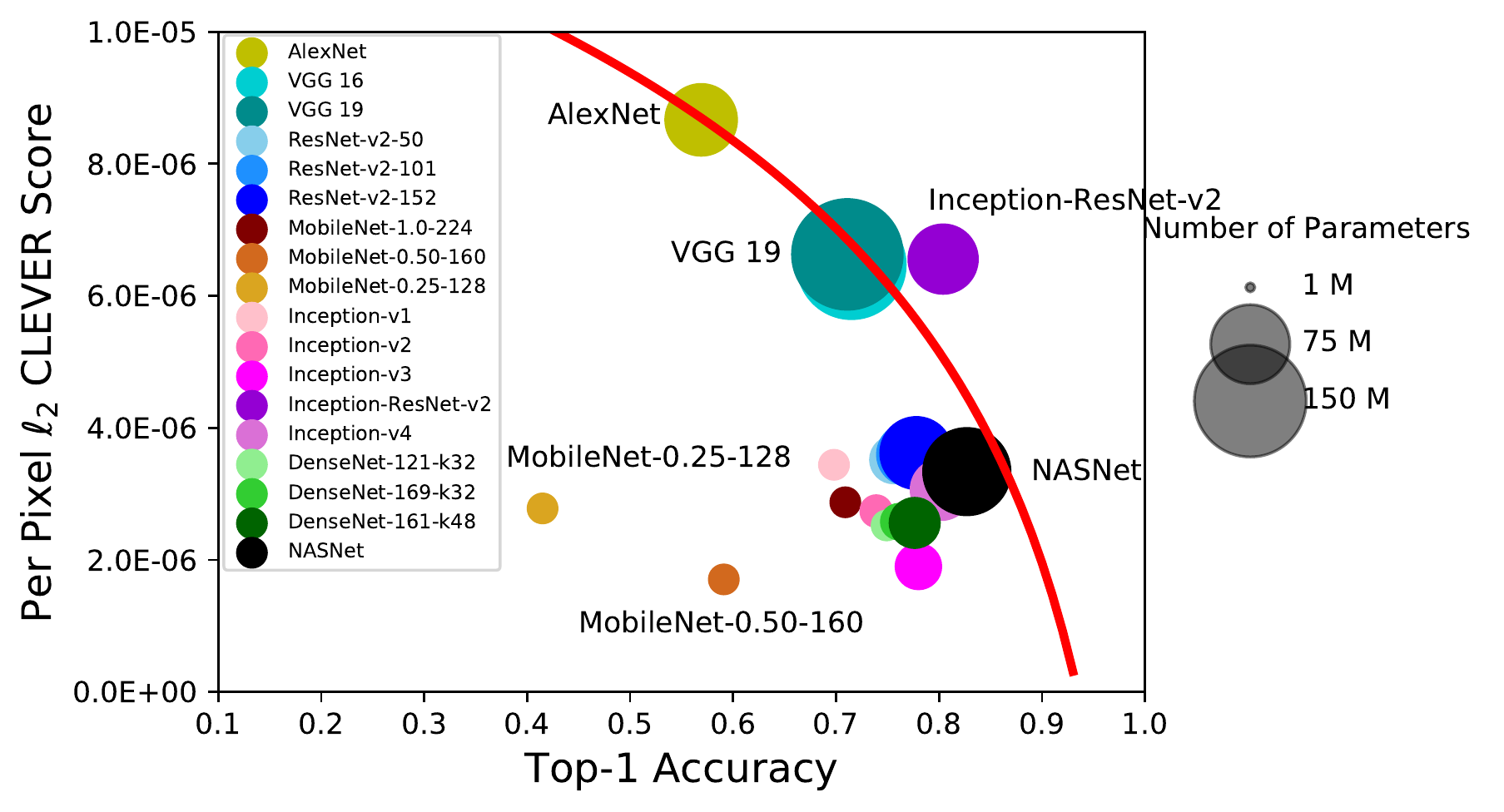}\\
    (c) Fitted Pareto frontier of $\ell_2$ CLEVER score vs. top-1 accuracy:\\ $\ell_2\text{ score}=[4.6\cdot\ln(1-\text{acc})+12.5]\times 10^{-6}$\\
    \end{tabular}
    \caption{Robustness vs. classification accuracy plots of I-FGSM attack~\cite{DBLP:journals/corr/KurakinGB16a}, C\&W attack~\cite{DBLP:conf/sp/Carlini017} and CLEVER~\cite{weng2018evaluating} score on random targets over 18 ImageNet models.}\label{fig:accuracy-robustness}
\end{figure*}

For C\&W and EAD-L1 attacks, increasing the confidence $\kappa$ can significantly make the attack harder to find a feasible adversarial example. A larger $\kappa$ usually makes the adversarial distortion more universal and improves transferability (as we will show shortly), but at the expense of decreasing the success rate and increasing the distortion. However, we find that the attack success rate with large $\kappa$ \textit{cannot be used as a robustness measure}, as it is not aligned with the $\ell_p$ norm of adversarial distortions. For example, for MobileNet-0.50-160, when $\kappa = 40$, the success rate is close to 0, but in Figure~\ref{fig:accuracy-robustness} we show that it is one of the most vulnerable networks. The reason is that the range of the logits output can be different for each network, so the difficulty of finding a fixed logit gap $\kappa$ is different on each network, and is not related to its intrinsic robustness.

We defer the results for targeted attack with the \emph{least likely} target label to the Supplementary section because the conclusions made are similar.

\begin{figure}[!htb]
\centering
	\includegraphics[width = 4.5in]{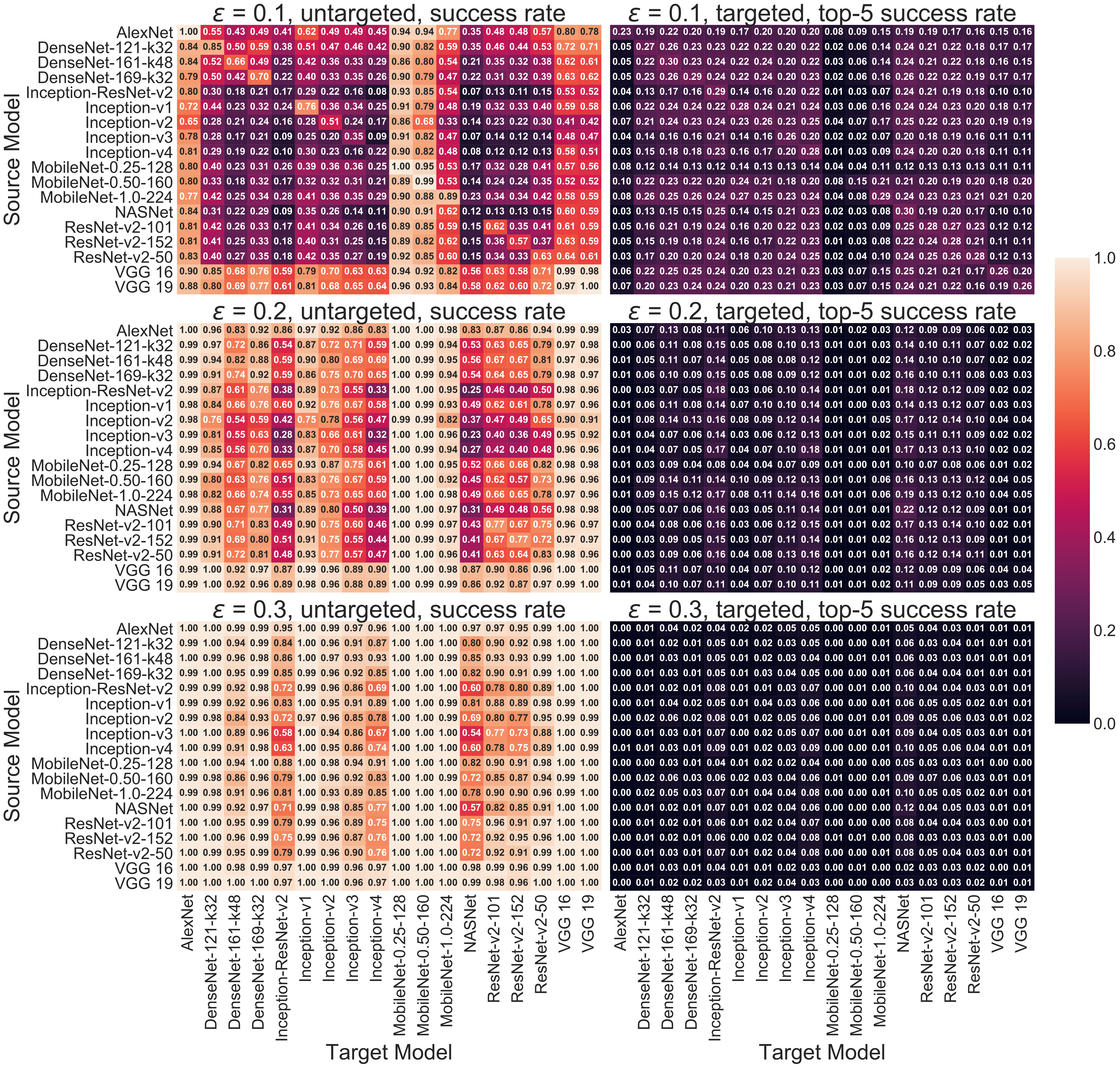}
	\caption{Transferability of FGSM attack over 18 ImageNet models.}\label{fig:fgsm_transferability}
\end{figure}

\begin{figure}[!htb]
\centering
	\includegraphics[width = 4.5in]{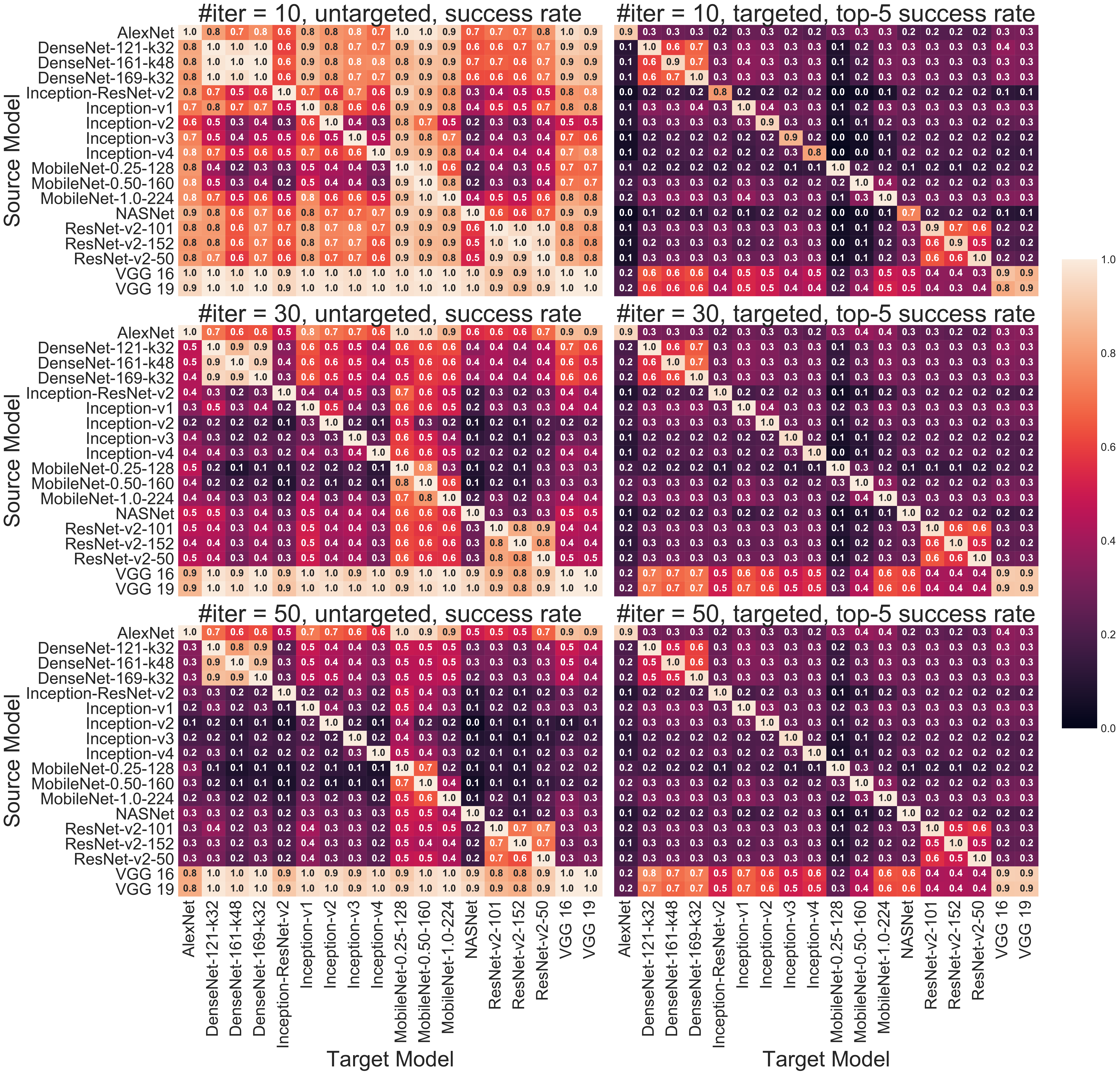}
	\caption{Transferability of I-FGSM attack over 18 ImageNet models, $\epsilon = 0.3$. }\label{fig:iter_fgsm_transferability}
\end{figure}


\begin{figure}[!htb]
\centering
	\includegraphics[width = 4.5in]{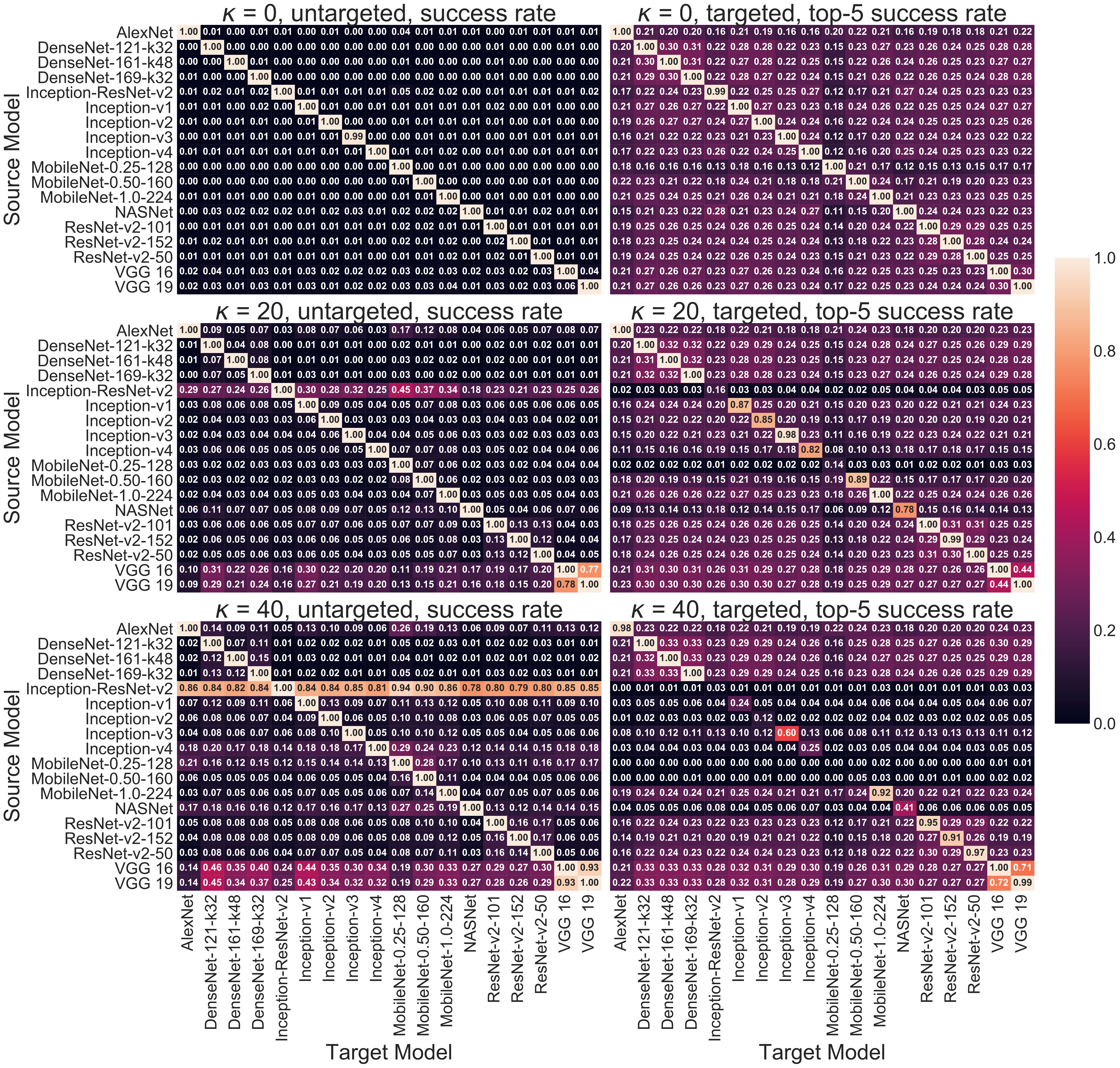}
	\caption{The transferability of EAD-L1 attack over 18 ImageNet models. }\label{fig:eadl1_transferability}
\end{figure}


\subsection{Linear Scaling Law in Robustness v.s. Accuracy }
Here we study the empirical relation between robustness and accuracy of different ImageNet models, where the robustness is evaluated in terms of the $\ell_\infty$ and $\ell_2$ distortion metrics from successful I-FGSM and C\&W attacks respectively, or $\ell_2$ CLEVER scores. In our experiments the attack success rates of these attacks are nearly 100\% for each model. The scatter plots of distortions/scores v.s. top-1 prediction accuracy are displayed in Figure~\ref{fig:accuracy-robustness}. We define the classification error as 1 minus top-1 accuracy (denoted as $1 - \text{acc}$). By regressing the distortion metric with respect to the  classification error of networks on the Pareto frontier of robustness-accuracy distribution (i.e., AlexNet, VGG~16, VGG~19, ResNet\_v2\_152, Inception\_ResNet\_v2 and NASNet), we find that 
the distortion scales linearly with the logarithm of classification error. That is, the distortion and classification error has the following relation: $\textnormal{distortion} = a+b \cdot \log \textnormal{(classification-error)}$. The fitted parameters of $a$ and $b$ are given in the captions of Figure~\ref{fig:accuracy-robustness}. Take I-FGSM attack as an example, the linear scaling law suggests that to reduce the classification error by a half, the $\ell_\infty$ distortion of the resulting network will be expected to reduce by approximately $0.02$, which is roughly $60\%$ of the AlexNet distortion. Following this trend, if we naively pursue a model with low test error, the model robustness may suffer. Thus,  when designing new networks for ImageNet, we suggest to evaluate the model's accuracy-robustness tradeoff by comparing it to the disclosed Pareto frontier.  


\subsection{Robustness of Different Model Sizes and Architectures}

We find that model architecture is a more important factor to model robustness than the model size. Each family of networks exhibits a similar level of robustness, despite different depths and model sizes. For example, AlexNet has about 60 million parameters but its robustness is the best; on the other hand, Mobilenet-0.50-160 has only 1.5 million parameters but is more vulnerable to adversarial attacks in all metrics.

We also observe that, within the same family, for DenseNet, ResNet and Inception, models with \textit{deeper architecture yields a slight improvement} of the robustness in terms of the $\ell_\infty$ distortion metric. This might provide new insights for designing robust networks and further improve the Pareto frontier. This result also echoes with~\cite{madry2017towards}, where the authors use a larger model to increase the $\ell_\infty$ robustness of a CNN based MNIST model.

\subsection{Transferability of Adversarial Examples}






Figures~\ref{fig:fgsm_transferability}, \ref{fig:iter_fgsm_transferability} and~\ref{fig:eadl1_transferability} show the transferability heatmaps of FGSM, I-FGSM and EAD-L1 over all 18 models (306 pairs in total). The value in the $i$-th row and $j$-th column of each heatmap matrix is the proportion of the adversarial examples successfully transferred to target model $j$ out of all adversarial examples generated by source model $i$ (including both successful and failed attacks on the source model). Specifically, the values on the diagonal of the heatmap are the attack success rate of the corresponding model. For each model, we generate adversarial images using the aforementioned attacks and pass them to the target model to perform black-box untargeted and targeted transfer attacks. To evaluate each model, we use the success rate for evaluating the untargeted transfer attacks and the top-5 matching rate for evaluating targeted transfer attacks.

Note that not all models have the same input image dimension. We also find that simply resizing the adversarial examples can significantly decrease the transfer attack success rate \cite{athalye2017synthesizing}. To alleviate the disruptive effect of image resizing on adversarial perturbations, when transferring an adversarial image from a network with larger input dimension to a smaller dimension, we crop the image from the center; conversely, we add a white boarder to the image when the source network's input dimension is smaller.

Generally, the transferability of untargeted attacks is significantly higher than that of targeted attacks, as indicated in Figure~\ref{fig:fgsm_transferability}, \ref{fig:iter_fgsm_transferability} and~\ref{fig:eadl1_transferability}.
We highlighted some interesting findings in our experimental results:
\begin{enumerate}
\item In the \textit{untargeted} transfer attack setting, FGSM and I-FGSM have much higher transfer success rates than those in EAD-L1 (despiting using a large $\kappa$). Similar to the results in~\cite{chen2017ead}, we find that the transferability of C\&W is even worse than that of EAD-L1 and we defer the results to the supplement. The ranking of attacks on transferability in untargeted setting is given by
\begin{align*}
\text{FGSM} \succeq \text{I-FGSM} \succeq \text{EAD-L1} \succeq \text{C\&W}.
\end{align*}

\item Again in the untargeted transfer attack setting, for FGSM, a larger $\epsilon$ yields better transferability, while for I-FGSM, \textit{less iterations} yield better transferability. For untargeted EAD-L1 transfer attacks, a higher $\kappa$ value (confidence parameter) leads to better transferability, but it is still far behind I-FGSM.

\item Transferability of adversarial examples is sometimes \textit{asymmetric}; for example, in Figure~\ref{fig:iter_fgsm_transferability}, adversarial examples of VGG~16 are highly transferable to Inception-v2, but adversarial examples of Inception-v2 do not transfer very well to VGG.

\item We find that VGG~16 and VGG~19 models achieve \textit{significantly better transferability} than other models, in both targeted and untargeted setting, for all attacking methods, leading to the ``stripe patterns''. This means that adversarial examples generated from VGG models are empirically more transferable to other models.
This observation might be explained by the simple convolutional nature of VGG networks, which is the stem of all other networks. VGG models are thus a good starting point for mounting black-box transfer attacks. We also observe that the most transferable model family may vary with different attacks. 

\item Most recent networks have some unique features that might restrict adversarial examples' transferability to only within the same family. For example, as shown in Figure~\ref{fig:iter_fgsm_transferability}, when using I-FGSM in the untargeted transfer attack setting, for DenseNets, ResNets and VGG, transferability between different depths of the same architecture is \textit{close to 100\%}, but their transfer rates to other architectures can be much worse. This provides us an opportunity to reserve-engineer the internal architecture of a black-box model, by feeding it with adversarial examples crafted for a certain architecture and measure the attack success rates.

\end{enumerate}

\section{Conclusions}

In this paper, we present the largest scale to date study on adversarial examples in ImageNet models. We show comprehensive experimental results on 18 state-of-the-art ImageNet models using adversarial attack methods focusing on $\ell_1$, $\ell_2$ and $\ell_\infty$ norms and also an attack-agnostic robustness score, CLEVER. Our results show that there is a clear trade-off between accuracy and robustness, and a better performance in testing accuracy in general reduces robustness. Tested on the ImageNet dataset, 
we discover an empirical linear scaling law between distortion metrics and the logarithm of classification errors in representative models.
We conjecture that following this trend, naively pursuing high-accuracy models may come with the great risks of lacking robustness.
We also provide a thorough adversarial attack transferability analysis between 306 pairs of these networks and discuss the robustness implications on network architecture.

In this work, we focus on image classification. To the best of our knowledge, the scale and profound analysis on 18 ImageNet models have not been studied thoroughly in the previous literature. We believe our findings could also provide insights to robustness and adversarial examples in other computer vision tasks such as object detection~\cite{xie2017adversarial} and image captioning~\cite{chen2018attacking}, since these tasks often use the same pre-trained image classifiers studied in this paper for feature extraction. 

\section{Supplementary Materials}
{
\let\section\subsection
\let\subsection\subsubsection
\section{Experiments on Images Correctly Classified by All Models}
To further validate our robustness analysis, we conducted another experiment by taking the subset of images (327 images in total) that are correctly classified by \textit{all} of 18 examined ImageNet models and show their accuracy-vs-robustness figures on C\&W and I-FGSM targeted attacks in Figure~\ref{fig:accuracy-robustness-targeted-random-labels-inter}. The trends and conclusions are consistent with our reported main results. 

\begin{figure*}[!h]
\centering
\begin{tabular}{c}\label{Fig:0}
	\includegraphics[width = \textwidth]{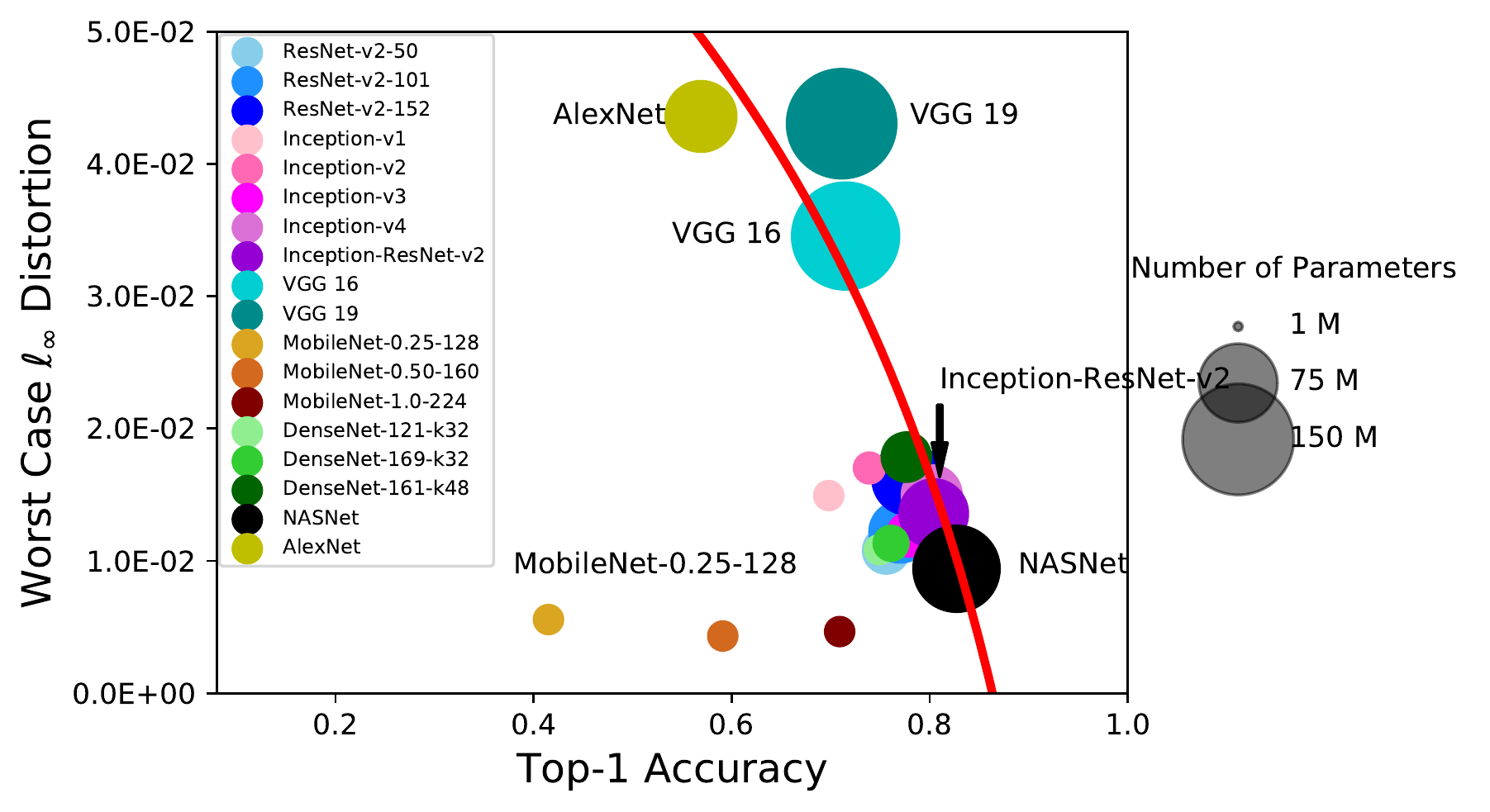}\\
	(a) Fitted Pareto frontier of $\ell_\infty$ distortion (I-FGSM attack) vs. top-1 accuracy: \\$\ell_\infty\text{ dist}=[4.3\cdot\ln(1-\text{acc})+8.5]\times 10^{-2}$\\
    \includegraphics[width = \textwidth]{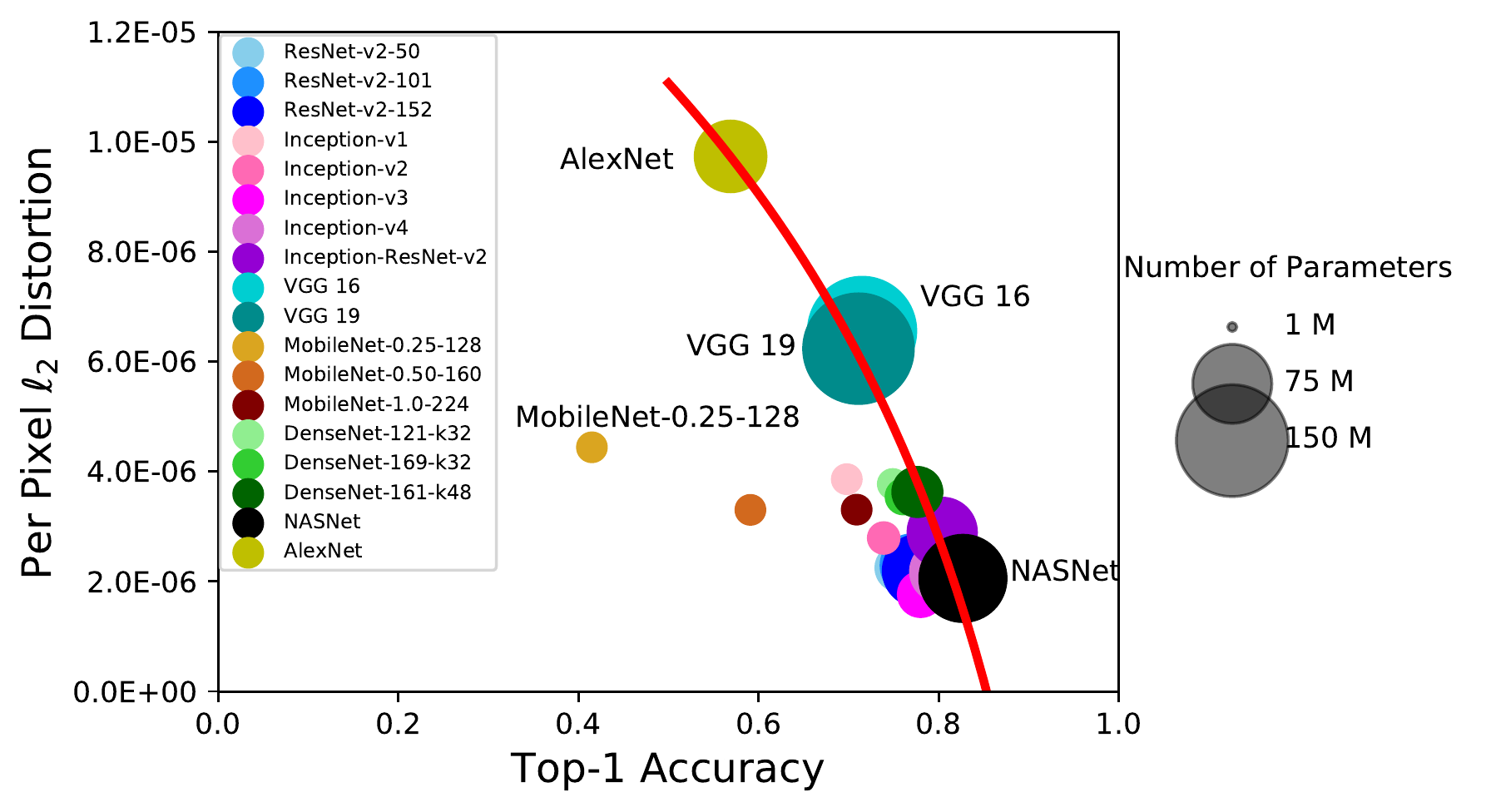}\\
	(b) Fitted Pareto frontier of $\ell_2$ distortion (C\&W attack) vs. top-1 accuracy:\\ $\ell_2\text{ dist}=[9.0\cdot\ln(1-\text{acc})+17.3]\times 10^{-6}$\\
    \end{tabular}
    \caption{Robustness vs. classification accuracy plots of I-FGSM attack~\cite{DBLP:journals/corr/KurakinGB16a}, C\&W attack~\cite{DBLP:conf/sp/Carlini017} on random targets on 18 ImageNet models based on 327 images correctly classified by all models.}\label{fig:accuracy-robustness-targeted-random-labels-inter}
\end{figure*}

\begin{figure*}[!h]
\centering
\begin{tabular}{c}\label{Fig:1}
	\includegraphics[width = 0.95\textwidth]{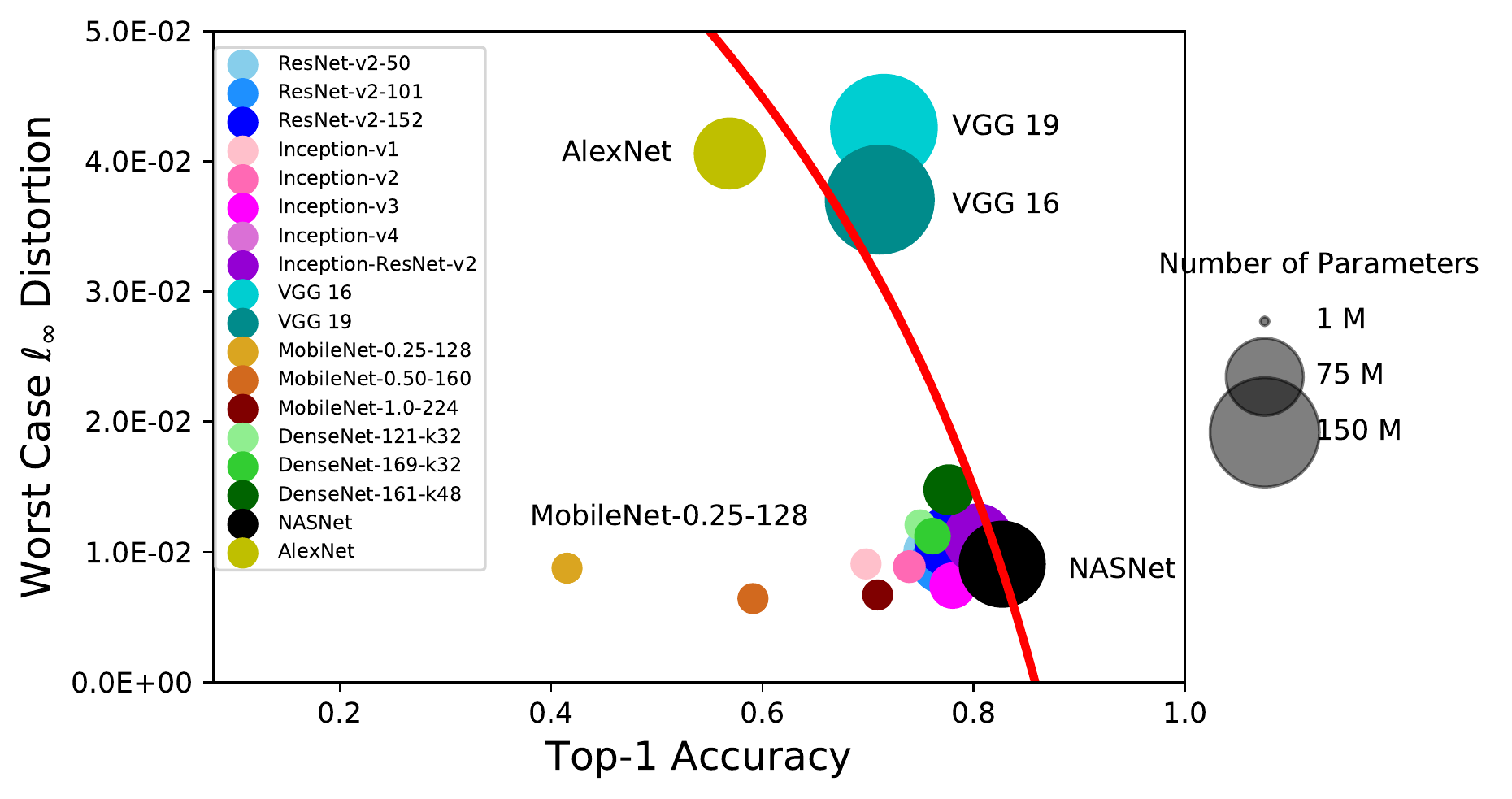}\\
	(a) Fitted Pareto frontier of $\ell_\infty$ distortion (I-FGSM attack) vs. top-1 accuracy: \\$\ell_\infty\text{ dist}=[4.3\cdot\ln(1-\text{acc})+8.5]\times 10^{-2}$\\
    \includegraphics[width = 0.95\textwidth]{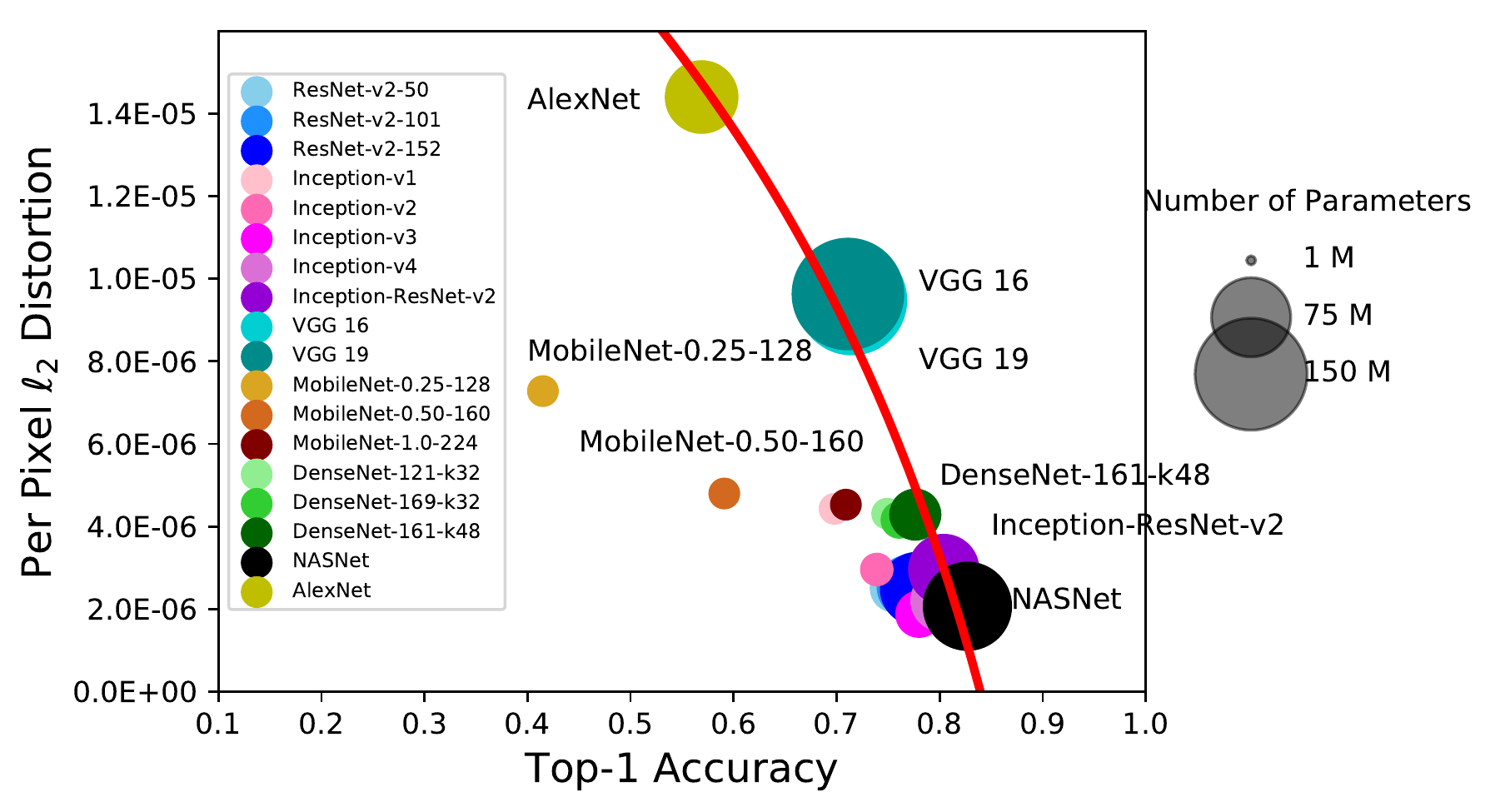}\\
	(b) Fitted Pareto frontier of $\ell_2$ distortion (C\&W attack) vs. top-1 accuracy:\\ $\ell_2\text{ dist}=[1.5\cdot\ln(1-\text{acc})+2.7]\times 10^{-5}$\\
    \end{tabular}
    \caption{Robustness vs. classification accuracy plots of I-FGSM attack~\cite{DBLP:journals/corr/KurakinGB16a}, C\&W attack~\cite{DBLP:conf/sp/Carlini017} on least likely targets on 18 ImageNet models.}\label{fig:accuracy-robustness-targeted-least-likely-labels}
\end{figure*}

\section{Robustness vs. Accuracy of Least-Likely Attacks}
In this section, we summarize the results of using the \textit{least-likely} label (the class with the smallest probability of the original image) as the target class. Figure~\ref{fig:accuracy-robustness-targeted-least-likely-labels} (a) and (b) show the distortions of adversarial examples found by I-FGSM and C\&W attacks, respectively. 
Although the least-likely label attack is even more challenging, both I-FGSM and C\&W algorithms can still achieve a close to 100\% success rate. Similar to Figure 2 of the main text, Figure~\ref{fig:accuracy-robustness-targeted-least-likely-labels} clearly shows an 
accuracy v.s. robustness trade-off for models on the Pareto frontier, e.g., AlexNet is the most robust network while the model with the highest accuracy (NASNet) is most prone to adversarial attacks. 
Likewise, we fit the Pareto frontier and still observe a similar log-linear scaling law. 

\section{The Transferability of C\&W Attack}

In this section, we show the transferability of C\&W attack in Figure~\ref{fig:cw_transferability}, \ref{fig:iter_fgsm_transferability_eps_0.1} and \ref{fig:iter_fgsm_transferability_eps_0.2}. Comparing with I-FGSM and EAD-L1 attacks, C\&W attack using $\ell_2$ norm yields a much worse transferability success rate. Increasing the confidence parameter $\kappa$ can slightly increase its transferability, but is still worse than that of I-FGSM and EAD. On the other hand, increasing $\kappa$ reduces C\&W attack's success rates, as we have shown in Figure 1 of the main text.  I-FGSM has much better transferability than EAD-L1 and C\&W attacks.  From Figure~\ref{fig:iter_fgsm_transferability_eps_0.1}, \ref{fig:iter_fgsm_transferability_eps_0.2} and Figure 4 in Section 3.4, we can see that the transferability increases as $\epsilon$ grows.

\section{More Experiments on the Transferability of I-FGSM Attack}
In this section, we show more experimental results on I-FGSM
attack with different $\epsilon$ values. Figures~\ref{fig:iter_fgsm_transferability_eps_0.1} and ~\ref{fig:iter_fgsm_transferability_eps_0.2} demonstrate the transferability heatmaps of I-FGSM with $\epsilon=0.1$ and $\epsilon=0.2.$ Comparing these two heatmaps with Figure~4 in the main text (transferability of I-FGSM with $\epsilon=0.3$), we observe that: (i) I-FGSM's transferability improves when $\epsilon$ increases; (ii) \textit{less iterations} usually yield better transferability; (iii) transferability of untargeted attacks is significantly higher than that of targeted attacks; (iv) adversarial examples of VGG networks consistently transfer very well; and (v) adversarial examples are easier to be transfered between the models sharing a same architecture (e.g., ResNets and DenseNets) but different depths.


\begin{figure}[!htb]
	\includegraphics[width = 4.8in]{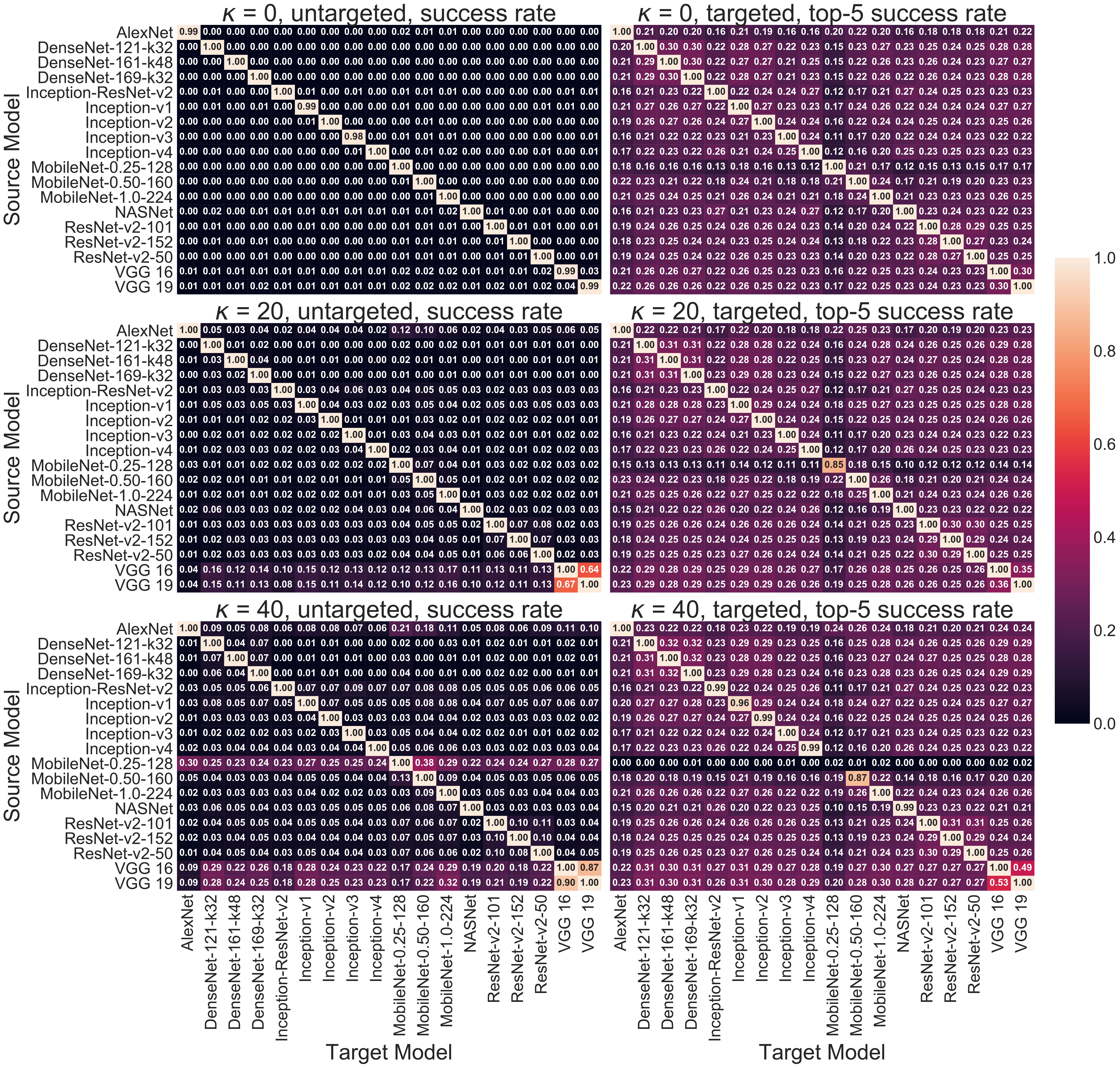}
	\caption{Transferability of C\&W attack over 18 ImageNet models.}\label{fig:cw_transferability}
\end{figure}

\begin{figure}[!htb]
	\includegraphics[width = 4.8in]{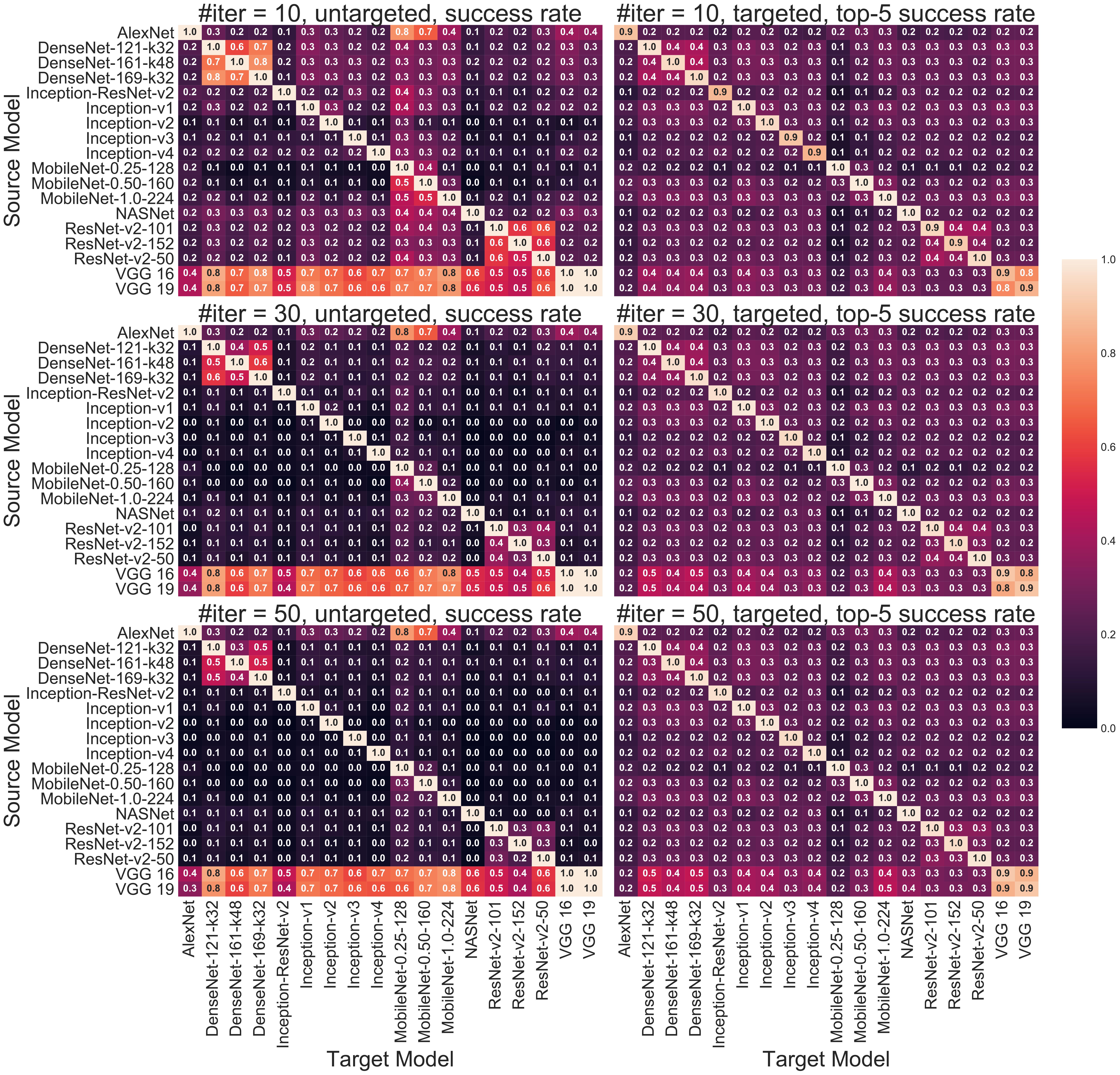}
	\caption{Transferability of I-FGSM attack over 18 ImageNet models with $\epsilon=0.1$. }\label{fig:iter_fgsm_transferability_eps_0.1}
\end{figure}

\begin{figure}[!htb]
	\includegraphics[width = 4.8in]{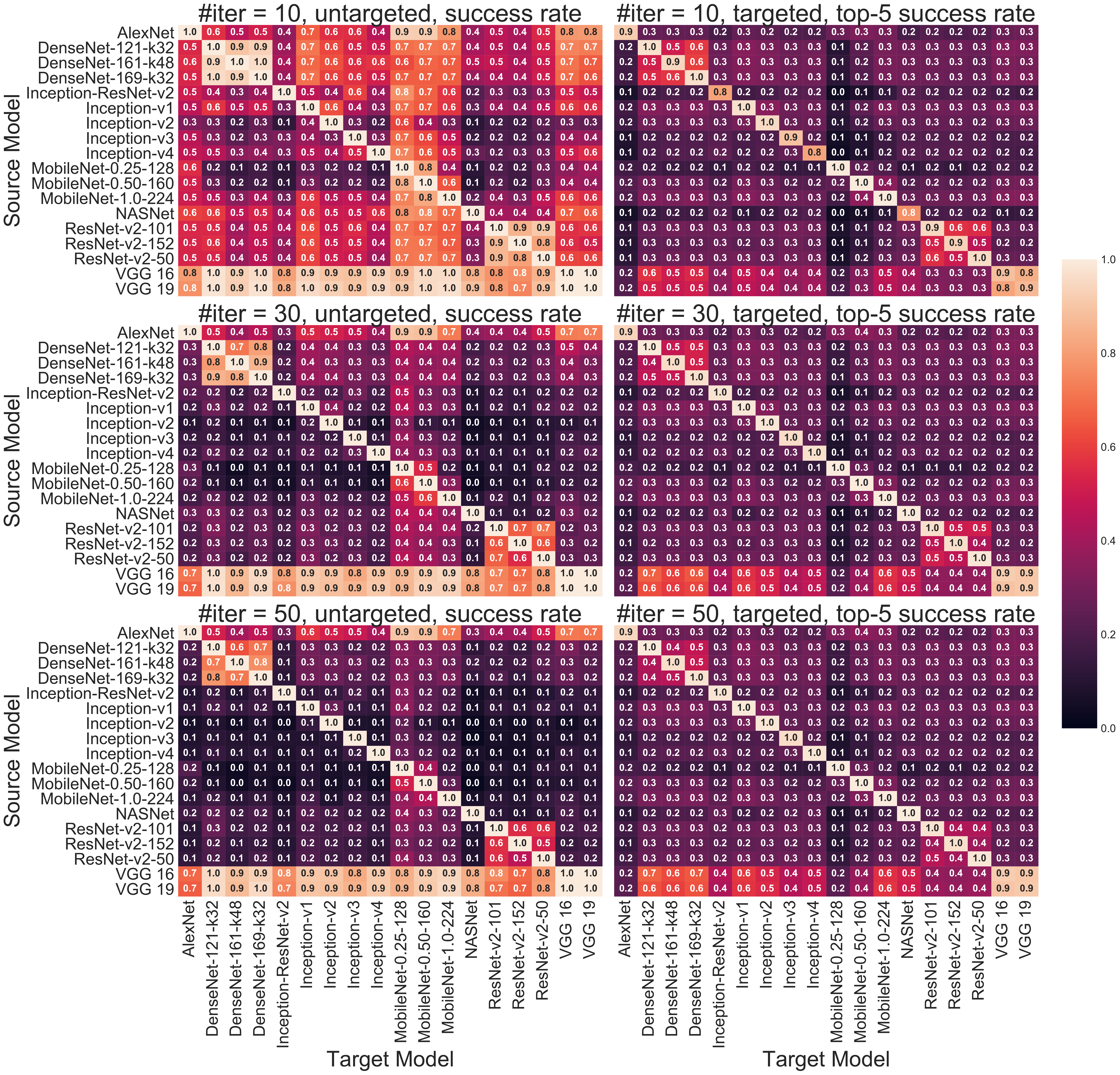}
	\caption{Transferability of I-FGSM attack over 18 ImageNet models with $\epsilon=0.2$. }\label{fig:iter_fgsm_transferability_eps_0.2}
\end{figure}

\section{Additional Remarks}

In~\cite{stock2017convnets} the authors also made a different conclusion on accuracy v.s. robustness. However, we believe our conclusion is not orthogonal to~\cite{stock2017convnets}, due to the apparent differences in the definition of ``robustness''. In~\cite{stock2017convnets}, the authors mainly explored the ``robustness'' (sensitivity) of class label semantics, where in the user study only 20 classes are selected and the I-FGSM attack with a fixed adversary strength is used. Each user is then asked to determine the adversarial label is ``relevant'' to the original  label or not, which is essentially a binarized class label relevance user study. The main message in~\cite{stock2017convnets} is that the inherent  correlations between image classes, if can be made more distinguishable (i.e., sensitivity as a strength), could be exploited towards building more accurate models. On the other hand, in our paper we used the standard $\ell_p$ ball perturbation in the pixel space as well as the attack success rates as the robustness measure on ImageNet with 1000 classes. In fact, the ``sensitivity'' issue has also been studied in~\cite{DBLP:journals/corr/KurakinGB16a} in terms of the ``label leaking'' effect. To ensure this effect has minimal impact when generating adversarial examples to evaluating the robustness of DNNs, the authors suggest including the attack results with ``least likely'' targets, which were included in this paper when drawing our conclusions.  

Images in ImageNet are organized according to the WordNet~\cite{miller1995wordnet} hierarchy. To justify that least likely labels used in our experiments are indeed irrelevant to the original labels, we show their corresponding synsets' shortest path distances in the WordNet hierarchy in Figure~\ref{fig:shortest_distance}.  We use Inception-v1 as the model in the experiment.  Two labels of shortest path distances greater than 5 are considered irrelevant. In our case, this applies to 96.6\% of our least likely attacks and hence the vulnerability is not from the label sensitivity effect as studied in~\cite{stock2017convnets}.

\begin{figure}[!htb]
\centering
    \includegraphics[width = 0.7\textwidth]{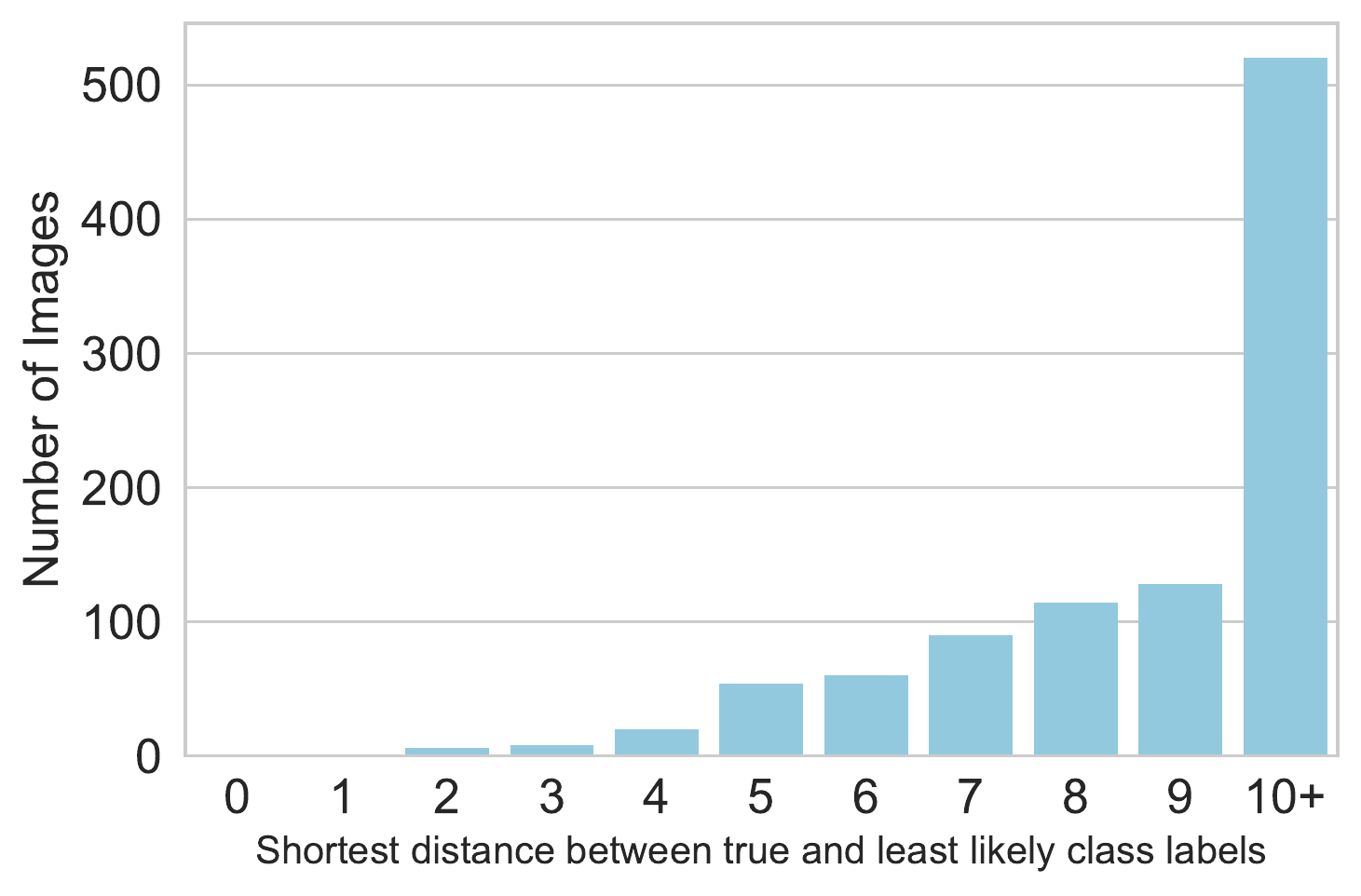}
    \label{fig:shortest_distance}
    \caption{The distribution of shortest distance to the true class from the least likely class in the WordNet hierarchy using 1,000 correctly classified ILSVRC'12 validation images.}\label{fig:shortest_path}
\end{figure}

In summary,~\cite{stock2017convnets}'s conclusion is that if one can increase the discriminative power against (semantically) similar classes, then the sensitivity in class labels could be a strength for model accuracy. Our conclusion is that more accurate network models appear to be less robust in terms of the required adversarial attack strength defined in $\ell_p$ ball.
Concurrent to this paper and similar to our conclusion, \cite{tsipras2018robustness} provides a concrete simple setting to demonstrate the trade-off between accuracy and robustness indeed provably exists, which also provides a technical explanation to our results.
We also note that our findings are consistent with the very recent paper~\cite{bubeck2018adversarial} that proves the difficulty of learning robust models against adversarial examples.

In light of~\cite{stock2017convnets}, our findings on accuracy-robustness trade-off could be explained by the increasing sensitivity in more accurate models -- these two robustness conclusions actually \textit{complement} each other, rather than being exclusive or contradictory. Specifically, increasing sensitivity aids in improved accuracy but might also make the model more vulnerable.  For example, increasing the sensitivity in classifying different dog species can improve the model accuracy, but may at the same time contribute to smaller adversarial perturbations. 
}

\bibliographystyle{splncs}
\bibliography{DL_Robustness}

\begin{thebibliography}{10}

\bibitem{krizhevsky2012imagenet}
Krizhevsky, A., Sutskever, I., Hinton, G.E.:
\newblock Imagenet classification with deep convolutional neural networks.
\newblock In: Advances in neural information processing systems (NIPS). (2012)
  1097--1105

\bibitem{DBLP:journals/ijcv/RussakovskyDSKS15}
Russakovsky, O., Deng, J., Su, H., Krause, J., Satheesh, S., Ma, S., Huang, Z.,
  Karpathy, A., Khosla, A., Bernstein, M.S., Berg, A.C., Li, F.:
\newblock Imagenet large scale visual recognition challenge.
\newblock International Journal of Computer Vision \textbf{115}(3) (2015)
  211--252

\bibitem{DBLP:journals/corr/GoodfellowSS14}
Goodfellow, I., Shlens, J., Szegedy, C.:
\newblock Explaining and harnessing adversarial examples.
\newblock In: International Conference on Learning Representations (ICLR).
  (2015)

\bibitem{xu2017can}
Xu, X., Chen, X., Liu, C., Rohrbach, A., Darell, T., Song, D.:
\newblock Fooling vision and language models despite localization and attention
  mechanism.
\newblock Proceedings of the Thirtieth IEEE/CVF Conference on Computer Vision
  and Pattern Recognition (CVPR) (2018)

\bibitem{chen2018attacking}
Chen, H., Zhang, H., Chen, P.Y., Yi, J., Hsieh, C.J.:
\newblock Attacking visual language grounding with adversarial examples: A case
  study on neural image captioning.
\newblock In: Proceedings of the 56th Annual Meeting of the Association for
  Computational Linguistics (Volume 1: Long Papers). Volume~1. (2018)
  2587--2597

\bibitem{metzen2017universal}
Metzen, J.H., Kumar, M.C., Brox, T., Fischer, V.:
\newblock Universal adversarial perturbations against semantic image
  segmentation.
\newblock stat \textbf{1050} (2017) ~19

\bibitem{cheng2018seq2sick}
Cheng, M., Yi, J., Zhang, H., Chen, P.Y., Hsieh, C.J.:
\newblock Seq2sick: Evaluating the robustness of sequence-to-sequence models
  with adversarial examples.
\newblock arXiv preprint arXiv:1803.01128 (2018)

\bibitem{carlini2018audio}
Carlini, N., Wagner, D.:
\newblock Audio adversarial examples: Targeted attacks on speech-to-text.
\newblock Deep Learning and Security Workshop (2018)

\bibitem{sun2018identify}
Sun, M., Tang, F., Yi, J., Wang, F., Zhou, J.:
\newblock Identify susceptible locations in medical records via adversarial
  attacks on deep predictive models.
\newblock In: Proceedings of the 24th ACM SIGKDD International Conference on
  Knowledge Discovery and Data Mining (KDD). (2018)  793--801

\bibitem{ijcai2018-543}
Xiao, C., Li, B., yan Zhu, J., He, W., Liu, M., Song, D.:
\newblock Generating adversarial examples with adversarial networks.
\newblock In: Proceedings of the Twenty-Seventh International Joint Conference
  on Artificial Intelligence, {IJCAI-18}, International Joint Conferences on
  Artificial Intelligence Organization (7 2018)  3905--3911

\bibitem{xiao2018spatially}
Xiao, C., Zhu, J.Y., Li, B., He, W., Liu, M., Song, D.:
\newblock Spatially transformed adversarial examples.
\newblock In: International Conference on Learning Representations (ICLR).
  (2018)

\bibitem{DBLP:journals/corr/EvtimovEFKLPRS17}
Eykholt, K., Evtimov, I., Fernandes, E., Li, B., Rahmati, A., Xiao, C.,
  Prakash, A., Kohno, T., Song, D.:
\newblock Robust physical-world attacks on deep learning visual classification.
\newblock In: Proceedings of the IEEE Conference on Computer Vision and Pattern
  Recognition. (2018)  1625--1634

\bibitem{szegedy2013intriguing}
Szegedy, C., Zaremba, W., Sutskever, I., Bruna, J., Erhan, D., Goodfellow, I.,
  Fergus, R.:
\newblock Intriguing properties of neural networks.
\newblock In: International Conference on Learning Representations (ICLR).
  (2014)

\bibitem{hein2017formal}
Hein, M., Andriushchenko, M.:
\newblock Formal guarantees on the robustness of a classifier against
  adversarial manipulation.
\newblock In: Advances in Neural Information Processing Systems 30: Annual
  Conference on Neural Information Processing Systems (NIPS). (2017)
  2263--2273

\bibitem{weng2018evaluating}
Weng, T.W., Zhang, H., Chen, P.Y., Yi, J., Su, D., Gao, Y., Hsieh, C.J.,
  Daniel, L.:
\newblock Evaluating the robustness of neural networks: An extreme value theory
  approach.
\newblock In: International Conference on Learning Representations (ICLR).
  (2018)

\bibitem{weng2018towards}
Weng, T.W., Zhang, H., Chen, H., Song, Z., Hsieh, C.J., Boning, D., Dhillon,
  I.S., Daniel, L.:
\newblock Towards fast computation of certified robustness for relu networks.
\newblock Proceedings of the 35th International Conference on Machine Learning
  {(ICML)} (2018)

\bibitem{stock2017convnets}
Stock, P., Cisse, M.:
\newblock Convnets and imagenet beyond accuracy: Explanations, bias detection,
  adversarial examples and model criticism.
\newblock arXiv preprint arXiv:1711.11443 (2017)

\bibitem{simonyan2014very}
Simonyan, K., Zisserman, A.:
\newblock Very deep convolutional networks for large-scale image recognition.
\newblock In: International Conference on Learning Representations (ICLR).
  (2015)

\bibitem{DBLP:conf/cvpr/SzegedyLJSRAEVR15}
Szegedy, C., Liu, W., Jia, Y., Sermanet, P., Reed, S.E., Anguelov, D., Erhan,
  D., Vanhoucke, V., Rabinovich, A.:
\newblock Going deeper with convolutions.
\newblock In: {IEEE} Conference on Computer Vision and Pattern Recognition,
  {CVPR} 2015, Boston, MA, USA, June 7-12, 2015. (2015)  1--9

\bibitem{DBLP:conf/cvpr/HeZRS16}
He, K., Zhang, X., Ren, S., Sun, J.:
\newblock Deep residual learning for image recognition.
\newblock In: 2016 {IEEE} Conference on Computer Vision and Pattern
  Recognition, {CVPR} 2016, Las Vegas, NV, USA, June 27-30, 2016. (2016)
  770--778

\bibitem{DBLP:journals/corr/HuangLW16a}
Huang, G., Liu, Z., van~der Maaten, L., Weinberger, K.Q.:
\newblock Densely connected convolutional networks.
\newblock In: 2017 {IEEE} Conference on Computer Vision and Pattern Recognition
  (CVPR). (2017)

\bibitem{DBLP:journals/corr/HowardZCKWWAA17}
Howard, A.G., Zhu, M., Chen, B., Kalenichenko, D., Wang, W., Weyand, T.,
  Andreetto, M., Adam, H.:
\newblock Mobilenets: Efficient convolutional neural networks for mobile vision
  applications.
\newblock CoRR \textbf{abs/1704.04861} (2017)

\bibitem{zoph2017learning}
Zoph, B., Vasudevan, V., Shlens, J., Le, Q.V.:
\newblock Learning transferable architectures for scalable image recognition.
\newblock In: 2018 {IEEE} Conference on Computer Vision and Pattern Recognition
  (CVPR). (2018)

\bibitem{lin2013network}
Lin, M., Chen, Q., Yan, S.:
\newblock Network in network.
\newblock In: International Conference on Learning Representations, ICLR
  (ICLR). (2014)

\bibitem{DBLP:conf/icml/IoffeS15}
Ioffe, S., Szegedy, C.:
\newblock Batch normalization: Accelerating deep network training by reducing
  internal covariate shift.
\newblock In: Proceedings of the 32nd International Conference on Machine
  Learning, {ICML} 2015, Lille, France, 6-11 July 2015. (2015)  448--456

\bibitem{DBLP:conf/cvpr/SzegedyVISW16}
Szegedy, C., Vanhoucke, V., Ioffe, S., Shlens, J., Wojna, Z.:
\newblock Rethinking the inception architecture for computer vision.
\newblock In: 2016 {IEEE} Conference on Computer Vision and Pattern
  Recognition, {CVPR} 2016, Las Vegas, NV, USA, June 27-30, 2016. (2016)
  2818--2826

\bibitem{DBLP:conf/aaai/SzegedyIVA17}
Szegedy, C., Ioffe, S., Vanhoucke, V., Alemi, A.A.:
\newblock Inception-v4, inception-resnet and the impact of residual connections
  on learning.
\newblock In: Proceedings of the Thirty-First {AAAI} Conference on Artificial
  Intelligence, February 4-9, 2017, San Francisco, California, {USA.} (2017)
  4278--4284

\bibitem{zoph2016neural}
Zoph, B., Le, Q.V.:
\newblock Neural architecture search with reinforcement learning.
\newblock In: International Conference on Learning Representations (ICLR).
  (2017)

\bibitem{tensorflow_models}
Wu, N., Sivakumar, S., Guadarrama, S., Andersen, D.:
\newblock Tensorflow-{S}lim {I}mage {C}lassification {M}odel {L}ibrary.
\newblock Github https://github.com/tensorflow/models/tree/master/research/slim
  (2017)

\bibitem{he2016identity}
He, K., Zhang, X., Ren, S., Sun, J.:
\newblock Identity mappings in deep residual networks.
\newblock In: European Conference on Computer Vision (ECCV), Springer (2016)
  630--645

\bibitem{liu2016delving}
Liu, Y., Chen, X., Liu, C., Song, D.:
\newblock Delving into transferable adversarial examples and black-box attacks.
\newblock In: International Conference on Learning Representations (ICLR).
  (2017)

\bibitem{papernot2016transferability}
Papernot, N., McDaniel, P., Goodfellow, I.:
\newblock Transferability in machine learning: from phenomena to black-box
  attacks using adversarial samples.
\newblock arXiv preprint arXiv:1605.07277 (2016)

\bibitem{chen2017zoo}
Chen, P.Y., Zhang, H., Sharma, Y., Yi, J., Hsieh, C.J.:
\newblock Zoo: Zeroth order optimization based black-box attacks to deep neural
  networks without training substitute models.
\newblock In: Proceedings of the 10th ACM Workshop on Artificial Intelligence
  and Security, ACM (2017)  15--26

\bibitem{DBLP:journals/corr/abs-1805-11770}
Tu, C., Ting, P., Chen, P., Liu, S., Zhang, H., Yi, J., Hsieh, C., Cheng, S.:
\newblock Autozoom: Autoencoder-based zeroth order optimization method for
  attacking black-box neural networks.
\newblock CoRR \textbf{abs/1805.11770} (2018)

\bibitem{cheng2018query}
Cheng, M., Le, T., Chen, P.Y., Yi, J., Zhang, H., Hsieh, C.J.:
\newblock Query-efficient hard-label black-box attack: An optimization-based
  approach.
\newblock arXiv preprint arXiv:1807.04457 (2018)

\bibitem{tu2018autozoom}
Tu, C.C., Ting, P., Chen, P.Y., Liu, S., Zhang, H., Yi, J., Hsieh, C.J., Cheng,
  S.M.:
\newblock Autozoom: Autoencoder-based zeroth order optimization method for
  attacking black-box neural networks.
\newblock arXiv preprint arXiv:1805.11770 (2018)

\bibitem{DBLP:journals/corr/KurakinGB16a}
Kurakin, A., Goodfellow, I.J., Bengio, S.:
\newblock Adversarial machine learning at scale.
\newblock In: International Conference on Learning Representations (ICLR).
  (2017)

\bibitem{cisse2017parseval}
Cisse, M., Bojanowski, P., Grave, E., Dauphin, Y., Usunier, N.:
\newblock Parseval networks: Improving robustness to adversarial examples.
\newblock In: International Conference on Machine Learning (ICML). (2017)
  854--863

\bibitem{DBLP:conf/sp/Carlini017}
Carlini, N., Wagner, D.A.:
\newblock Towards evaluating the robustness of neural networks.
\newblock In: 2017 {IEEE} Symposium on Security and Privacy (Oakland) 2017, San
  Jose, CA, USA, May 22-26, 2017. (2017)  39--57

\bibitem{DBLP:journals/corr/CarliniW17}
Carlini, N., Wagner, D.:
\newblock Adversarial examples are not easily detected: Bypassing ten detection
  methods.
\newblock In: Proceedings of the 10th ACM Workshop on Artificial Intelligence
  and Security. AISec '17, New York, NY, USA, ACM (2017)  3--14

\bibitem{chen2017ead}
Chen, P.Y., Sharma, Y., Zhang, H., Yi, J., Hsieh, C.J.:
\newblock Ead: Elastic-net attacks to deep neural networks via adversarial
  examples.
\newblock AAAI (2018)

\bibitem{sharma2017breaking}
Sharma, Y., Chen, P.Y.:
\newblock Attacking the {Madry} defense model with ${L_1}$--based adversarial
  examples.
\newblock arXiv preprint arXiv:1710.10733 (2017)

\bibitem{lu2018limitation}
Lu, P.H., Chen, P.Y., Chen, K.C., Yu, C.M.:
\newblock On the limitation of magnet defense against ${L_1}$-based adversarial
  examples.
\newblock IEEE/IFIP DSN Workshop (2018)

\bibitem{lu2018limitation2}
Lu, P.H., Chen, P.Y., Yu, C.M.:
\newblock On the limitation of local intrinsic dimensionality for
  characterizing the subspaces of adversarial examples.
\newblock ICLR Workshop (2018)

\bibitem{deng2009imagenet}
Deng, J., Dong, W., Socher, R., Li, L.J., Li, K., Fei-Fei, L.:
\newblock Imagenet: A large-scale hierarchical image database.
\newblock In: Computer Vision and Pattern Recognition, 2009. CVPR 2009. IEEE
  Conference on, IEEE (2009)  248--255

\bibitem{krizhevsky2009learning}
Krizhevsky, A.:
\newblock Learning multiple layers of features from tiny images.
\newblock (2009)

\bibitem{DBLP:journals/nn/StallkampSSI12}
Stallkamp, J., Schlipsing, M., Salmen, J., Igel, C.:
\newblock Man vs. computer: Benchmarking machine learning algorithms for
  traffic sign recognition.
\newblock Neural Networks \textbf{32} (2012)  323--332

\bibitem{DBLP:conf/cvpr/Moosavi-Dezfooli16}
Moosavi{-}Dezfooli, S., Fawzi, A., Frossard, P.:
\newblock Deepfool: {A} simple and accurate method to fool deep neural
  networks.
\newblock In: 2016 {IEEE} Conference on Computer Vision and Pattern
  Recognition, {CVPR} 2016, Las Vegas, NV, USA, June 27-30, 2016. (2016)
  2574--2582

\bibitem{madry2017towards}
Madry, A., Makelov, A., Schmidt, L., Tsipras, D., Vladu, A.:
\newblock Towards deep learning models resistant to adversarial attacks.
\newblock In: International Conference on Learning Representations (ICLR).
  (2018)

\bibitem{athalye2017synthesizing}
Athalye, A., Engstrom, L., Ilyas, A., Kwok, K.:
\newblock Synthesizing robust adversarial examples.
\newblock 35th International Conference on Machine Learning (ICML) (2018)

\bibitem{xie2017adversarial}
Xie, C., Wang, J., Zhang, Z., Zhou, Y., Xie, L., Yuille, A.:
\newblock Adversarial examples for semantic segmentation and object detection.
\newblock In: International Conference on Computer Vision (ICCV), IEEE (2017)

\bibitem{miller1995wordnet}
Miller, G.A.:
\newblock Wordnet: a lexical database for english.
\newblock Communications of the ACM \textbf{38}(11) (1995)  39--41

\bibitem{tsipras2018robustness}
Tsipras, D., Santurkar, S., Engstrom, L., Turner, A., Madry, A.:
\newblock Robustness may be at odds with accuracy.
\newblock In: International Conference on Learning Representations (ICLR).
  (2019)

\bibitem{bubeck2018adversarial}
Bubeck, S., Price, E., Razenshteyn, I.:
\newblock Adversarial examples from computational constraints.
\newblock arXiv preprint arXiv:1805.10204 (2018)

\end{thebibliography}

\end{document}